\documentclass{article}
\usepackage{multirow}
\usepackage{dsfont}
\usepackage{amsmath}
\usepackage{amsthm}
\usepackage{amsfonts}
\usepackage{algorithm}
\usepackage{algorithmicx}
\usepackage{algpseudocode}
\usepackage{authblk}

\usepackage{graphics}
\usepackage{graphicx}

\usepackage{tikz}
\usetikzlibrary{arrows}
\usepackage{bm}
\usepackage{array}

\DeclareMathOperator*{\minimize}{minimize}
\DeclareMathOperator*{\maximize}{maximize}
\DeclareMathOperator*{\argmax}{argmax}
\DeclareMathOperator*{\argmin}{argmin}
\DeclareMathOperator*{\diam}{diam}

\begin{document}

\newcommand{\bc}{\ensuremath{\mathbf{c}}}
\newcommand{\bci}[1]{\ensuremath{\mathbf{c}_{#1}}}
\newcommand{\bz}{\ensuremath{\mathbf{z}}}
\newcommand{\be}{\ensuremath{\mathbf{e}}}
\newcommand{\bei}[1]{\ensuremath{\mathbf{e}}(#1)}
\newcommand{\bzi}[1]{\ensuremath{\mathbf{z}_{#1}}}
\newcommand{\bx}{\ensuremath{\mathbf{x}}}
\newcommand{\bw}{\ensuremath{\mathbf{w}}}
\newcommand{\bp}{\ensuremath{\mathbf{p}}}
\newcommand{\bxi}[1]{\ensuremath{\mathbf{x}_{#1}}}
\newcommand{\cBi}[1]{\ensuremath{\mathcal{B}_{#1}}}
\newcommand{\cB}{\ensuremath{\mathcal{B}}}
\newcommand{\cX}{\ensuremath{\mathcal{X}}}
\newcommand{\cZ}{\ensuremath{\mathcal{Z}}}
\newcommand{\cc}{(\ensuremath{\mathbf{c}_{1}-\mathbf{c}_{2})}}
\newcommand{\ncc}{\ensuremath{\|\mathbf{c}_{1}-\mathbf{c}_{2}\|^{2}}}
\newcommand{\norm}[1]{\ensuremath{\|#1\|^{2}}}%
\newcommand{\bbR}{\ensuremath{\mathbb{R}}}

\renewcommand{\algorithmicrequire}{\textbf{Input:}}

\newtheorem{definition}{Definition}
\newtheorem{proposition}{Proposition}
\newtheorem{theorem}{Theorem}

\title{Training Support Vector Machines Using Frank-Wolfe Optimization Methods}

\author[1]{Emanuele Frandi}
\author[2]{Ricardo \~Nanculef}
\author[3]{Maria Grazia Gasparo}
\author[4]{Stefano Lodi}
\author[5]{Claudio Sartori}
\affil[1]{\small Dept. of Science and High Technology, University of Insubria, Italy \newline \texttt{emanuele.frandi@uninsubria.it}}
\affil[2]{\small Dept. of Informatics, Federico Santa Mar\'ia University, Chile \newline \texttt{jnancu@inf.utfsm.cl}}
\affil[3]{\small Dept. of Energetics, University of Florence, Italy \newline \texttt{mariagrazia.gasparo@unifi.it}}
\affil[4,5]{\small Dept. of Electronics, Computer Science and Systems, University of Bologna, Italy \newline \texttt{\{stefano.lodi,claudio.sartori\}@unibo.it}}
\date{}

\maketitle
\thispagestyle{empty}

\begin{abstract}
Training a Support Vector Machine (SVM) requires the solution of a quadratic programming problem (QP) whose computational complexity becomes prohibitively expensive for large scale datasets. Traditional optimization methods cannot be directly applied in these cases, mainly due to memory restrictions.

By adopting a slightly different objective function and under mild conditions on the kernel used within the model, efficient algorithms to train SVMs
have been devised under the name of Core Vector Machines (CVMs). This framework exploits the equivalence of the resulting learning problem
with the task of building a Minimal Enclosing Ball (MEB) problem in a feature space, where data is implicitly embedded by a kernel function. 

In this paper, we improve on the CVM approach by proposing two novel methods to build SVMs based on the Frank-Wolfe algorithm, recently revisited as a fast method to approximate the solution of a MEB problem. In contrast to CVMs, our algorithms do not require to compute the solutions of a sequence of increasingly complex QPs and are defined by using only analytic optimization steps. Experiments on a large collection of datasets show that our methods scale better than CVMs in most cases, sometimes at the price of a slightly lower accuracy. As CVMs, the proposed methods can be easily extended to machine learning problems other than binary classification. However, effective classifiers are also obtained using kernels which do not satisfy the condition required by CVMs and can thus be used for a wider set of problems.
\end{abstract}

\section{Introduction}
Support Vector Machines (SVMs) are currently one of the most effective methods to approach classification and other machine learning problems, improving on more traditional techniques like decision trees and neural networks in a number of applications \cite{Hofmann2008,Scholkopf01Kernels}. SVMs are defined by optimizing a regularized risk functional on the training data, which in most cases leads to classifiers with an outstanding generalization performance \cite{Vapnik95TheNature,Scholkopf01Kernels}. This optimization problem is usually formulated as a large convex
quadratic programming problem (QP), for which a naive implementation requires $O(m^2)$ space and $O(m^3)$ time in the number of examples
$m$, complexities that are prohibitively expensive for large scale problems \cite{Scholkopf01Kernels,coreSVMs05tsang}. Major research efforts have been hence directed towards scaling up SVM algorithms to large datasets.

Due to the typically dense structure of the hessian matrices involved in the QP, traditional optimization methods cannot be
directly applied to train an SVM on large datasets. The problem is usually addressed using an \textit{active set method} where at each
iteration only a small number of variables are allowed to change \cite{active-set-implementation06scheinberg,joachims99making-large-scale-svms-practical,platt99smo-seminal}. In non-linear SVMs problems, this is essentially equivalent to selecting a subset of training examples called a \emph{working set} \cite{Vapnik95TheNature}. The most prominent example in this category of methods is Sequential Minimal Optimization (SMO), where only two variables are selected for optimization each time \cite{smo-second-order05fan,platt99smo-seminal}. The main disadvantage of these methods is that they generally exhibit a slow local rate of convergence, that is, the closer one gets to a solution, the more slowly one approaches that solution. Moreover, performance results are in practice very sensitive to the size of the active set, the way to select the active variables and other implementation details like the caching strategy used to avoid repetitive computations of the kernel function on which the model is based \cite{active-set-implementation06scheinberg} Other attempts to scale up SVM methods consist in adapting \textit{interior point methods} to some classes of the SVM
QP.\cite{low-rank-representations02fine}. For large-scale problems however the resulting rank of the kernel matrix can still be too
high to be handled efficiently \cite{coreSVMs05tsang}. The reformulation of the SVM objective function as in
\cite{relaxation-of-the-QP03fung-mangasarian}, the use of sampling methods to reduce the number of variables in the problem as
in \cite{reducedSVMs07lee} and \cite{randomizedSVM08}, and the combination of small SVMs using ensemble methods as in
\cite{boosting-svms00Pavlov} have also been explored.

Looking for more efficient methods, in \cite{coreSVMs05tsang} a new approach was proposed: the task of learning the classifier from data can be transformed to the problem of computing a \textit{minimal enclosing ball} (MEB), that is, the ball of smallest radius containing a set of points. This equivalence is obtained by adopting a slightly different penalty term in the objective function and imposing some mild conditions on the kernel used by the SVM. Recent advances in computational geometry have demonstrated that there are algorithms capable of approximating a MEB with any degree of accuracy $\epsilon$ in $O(1/\epsilon)$ iterations independently of the number of points and the dimensionality of the space in which the ball is built \cite{coreSVMs05tsang}. Adopting one of these algorithms, Tsang and colleagues devised in \cite{coreSVMs05tsang} the Core Vector Machine (CVM), demonstrating that the new method compares favorably with most traditional SVM software, including for example software based on SMO \cite{smo-second-order05fan,platt99smo-seminal}.

CVMs start by solving the optimization problem on a small subset of data and then proceed iteratively. At each iteration the algorithm
looks for a point outside the approximation of the MEB obtained so far. If this point exists, it is added to the previous subset of data to define a larger optimization problem, which is solved to obtain a new approximation to the MEB. The process is repeated until no points outside the current approximating ball are found within a prescribed tolerance. CVMs hence need the resolution of a sequence of optimization problems of increasing complexity using an external numerical solver. In order to be efficient, the solver should be able to solve each problem from a warm-start and to avoid the full storage of the corresponding Gram matrix. Experiments in Ref. 37 employ to this end a variant of the second-order SMO proposed in \cite{smo-second-order05fan}.

In this paper, we study two novel algorithms that exploit the formalism of CVMs but do not need the resolution of a sequence of QPs. 
These algorithms are based on the Frank-Wolfe (FW) optimization framework, introduced in \cite{wolfe1954} and recently studied in
\cite{yildirim08} and \cite{clarkson08coresets} as a method to
approximate the solution of the MEB problem and other convex
optimization problems defined on the unit simplex. Both algorithms
can be used to obtain a solution arbitrarily close to the optimum, but
at the same time are considerably simpler than CVMs. The key idea is
to replace the nested optimization problem to be solved at each
iteration of the CVM approach by a linearization of the objective
function at the current feasible solution and an exact line search
in the direction obtained from the linearization. Consequently, each
iteration becomes fairly cheaper than a CVM iteration and
does not require any external numerical solver.

Similar to CVMs, both algorithms incrementally discover the examples which
become support vectors in the SVM model, looking for the optimal set
of weights in the process. However, the second of the proposed
algorithms is also endowed with the ability to explicitly remove examples
from the working set used at each iteration of the procedure and has
thus the potential to compute smaller models. On the theoretical
side, both algorithms are guaranteed to succeed in $O(1/\epsilon)$
iterations for an arbitrary $\epsilon > 0$. In addition, the second
algorithm exhibits an asymptotically linear rate of
convergence \cite{yildirim08}.

This research was originally motivated by the use of the MEB
framework and computational geometry optimization for the problem of
training an SVM. However, a major advantage of the proposed methods
over the CVM approach is the possibility to employ kernels which do
not satisfy the conditions required to obtain the equivalence
between the SVM and MEB optimization problems. For example, the
popular polynomial kernel does not allow the use of CVMs as a training
method. Since the optimal kernel for a given application cannot be
specified \emph{a priori}, the capability of a training method to
work with any valid kernel function is an important feature.
Adaptations of the CVM to handle more general kernels have been
recently proposed in \cite{coreSVMs-generalized06tsang} but,
in contrast, our algorithms can be used with any Mercer kernel
without changes to the theory or the implementation.

The effectiveness of the proposed methods is evaluated on several data classification sets, most of them already used to show the
improvements of CVMs over second-order SMO \cite{coreSVMs05tsang}. Our experimental results suggest that, as long as a minor loss in accuracy is acceptable, our algorithms  significantly improve the actual running times of this algorithm. Statistical tests are conducted to assess the significance of these conclusions. In addition, our experiments confirm that effective classifiers are also obtained with kernels that do not fulfill the conditions required by CVMs.

The article is organized as follows. Section $2$
presents a brief overview on SVMs and the way by which the problem
of computing an SVM can be treated as a MEB problem. Section $3$
describes the CVM approach. In Section $4$ we introduce the proposed
methods. Section $5$ presents the experimental setting and our
numerical results. Section $6$ closes the article with a discussion
of the main conclusions of this research.

\section{Support Vector Machines and the MEB Equivalence}

In this section we present an overview on Support Vector Machines (SVMs), and discuss the conditions under which the problem of building these models can be treated as a Minimal Enclosing Ball (MEB) problem in a feature space.

\subsection{The Pattern Classification Problem}

Consider a set of training data $S= \{\mathbf{x}_{i}\}$ with $\mathbf{x}_{i} \in
\mathcal{X}$, $i \in \mathcal{I} = \{1,\ldots,m\}$. The set $\mathcal{X}$, often coinciding with $\mathbb{R}^n$, is called the \emph{input space}, and each instance is
associated with a given category in the set $C=\{c_{1},c_{2},
\ldots, c_{K}\}$. A pattern classification problem consists of
inferring from $S$ a prediction mechanism $f: \mathcal{X}
\rightarrow C \in \mathcal{F}$, termed hypothesis, to associate new
instances $\mathbf{x} \in \mathcal{X}$ with the correct category.
When $K=2$ the problem described above is called binary
classification. This problem can be addressed by defining a set of
candidate models $\mathcal{F}$, a risk functional $R_l(S,f)$ assessing
the ability of $f$ to correctly predict the category of the
instances in $\mathcal{X}$, and a procedure $L$ by which a dataset
$S$ is mapped to a given hypothesis $f = L(S) \in \mathcal{F}$
achieving a low risk. In the context of machine learning, $L$ is
called the learning algorithm, $\mathcal{F}$ the hypothesis space
and $R_l(S,f)$ the induction principle \cite{Scholkopf01Kernels}.
%, and $R_l(S,f)$ the induction principle.

In the rest of this paper we focus on the problem of computing a
model designed for binary classification problems. The extension of
these models to handle multiple categories can be accomplished in several
ways. A possible approach corresponds to use several binary classifiers,
separately trained and joined into a multi-category decision
function. Well known approaches of this type are one-versus-the-rest
(OVR, see \cite{Vapnik95TheNature}), where one classifier is
trained to separate each class from the rest; one-versus-one
(OVO, see \cite{Ulrich99Pairwise}), where different binary SVMs are used
to separate each possible pair of classes; and DDAG, where one-versus-one classifiers are
organized in a directed acyclic graph decision structure \cite{Scholkopf01Kernels}. Previous
experiments with SVMs show that OVO frequently obtains a
better performance both in terms of accuracy and training time \cite{Hsu02ComparisonMultiClassSVMs}. 
Another type of extension consists in reformulating the optimization problem underlying the
method to directly address multiple categories. See \cite{Crammer01AlgImpl}, \cite{Lee04},
\cite{ADSVM} and \cite{MultiCore} for details about these
methods.

\subsection{Linear Classifiers and Kernels}

Support Vector Machines implement the decision mechanism by using
simple linear functions. Since in realistic problems the
configuration of the data can be highly non-linear, SVMs build a
linear model not in the original space $\mathcal{X}$, but in a
high-dimensional dot product \emph{feature space} $\mathcal{P}=
\overline{\mbox{lin}\left(\phi(\mathcal{X})\right)}$ where the
original data is embedded through the mapping $\bp = \phi(\bx)$ for each
$\mathbf{x} \in \mathcal{X}$. In
this space, it is expected that an accurate decision function can be
linearly represented. The feature space is related with $\mathcal{X}$ by
means of a so called \emph{kernel function} $k:\mathcal{X}  \times
\mathcal{X} \rightarrow \mathbb{R}$, which allows to compute dot
products in $\mathcal{P}$ directly from the input space. More precisely, for
each $\mathbf{x}_{i}$, $\mathbf{x}_{j}$ $\in \mathcal{X}$, we have
$\mathbf{p}_{i}^{T}\mathbf{p}_{j} = k(\mathbf{x}_{i},
\mathbf{x}_{j})$. The explicit computation of the
mapping $\phi$, which would be computationally infeasible, is thus avoided \cite{Scholkopf01Kernels}. For binary
classification problems, the most common approach is to associate a
positive label $y_i=+1$ to the examples of the first class, and a negative label $y_i=-1$ to the examples belonging
to the other class. This approach allows the use of real valued
hypotheses $h: \mathcal{P} \rightarrow \bbR$, whose output is passed through a
sign threshold to yield the classification label
$f(\bx)=\mbox{sgn}(h(\bp))=\mbox{sgn}(h(\phi(\bx)))$. Since $h(\bp)$
is a linear function in $\mathcal{P}$, the final prediction mechanism takes
the form
\begin{equation}
f(\mathbf{x}) = \mbox{sgn}\left(h(\phi(\mathbf{x}))\right) =
\mbox{sgn}\left( \mathbf{w}^{T}\phi(\mathbf{x}) + b \right)
\label{eq:decfunc} \, ,
\end{equation}
with $\mathbf{w} \in \mathcal{P}$ and $b \in \bbR$. This gives a
classification rule whose decision boundary $H = \{\bp:
\mathbf{w}^{T}\bp + b = 0\}$ is a hyperplane with normal vector
$\mathbf{w}$ and position parameter $b$.

\subsection{Large Margin Classifiers}

It should be noted that a decision function well predicting the
training data does not necessarily classify well unseen examples. Hence,
minimizing the training error (or \emph{empirical risk})
\begin{equation}
\sum_{i \in \mathcal{I}} \frac{1}{2} \left|1 - y_i f(\bx_i)\right| \, ,
\end{equation}
does not necessarily imply a small test error. The implementation of
an induction principle $R_l(S,f)$ guaranteeing a good classification
performance on new instances of the problem is addressed in SVMs by
building on the concept of \emph{margin} $\rho$. For a given training pair
$\bx_i,y_i$, the margin is defined as $\rho_{i}=\rho_f(\bx_{i},y_i)
= y_{i} h(\bx_{i}) = y_i \left( \mathbf{w}^{T}\phi(\mathbf{x}) + b
\right)$ and is expected to estimate how reliable the prediction of
the model on this pattern is. Note that the example $\mathbf{x}_{i}$
is misclassified if and only if $\rho_i < 0$. Note also that a large
margin of the pattern $\mathbf{x}_{i}$ suggests a more robust
decision with respect to changes in the parameters of the decision
function $f(\bx)$, which are to be estimated from the training
sample \cite{Scholkopf01Kernels}. The margin attained by a given
prediction mechanism on the full training set $S$ is defined as the
minimum margin over the whole sample, that is $\rho = \min_{i\in
\mathcal{I}}\rho_f(\bx_{i},y_i)$. This implements a measure of the
worst classification performance on the training set, since
$\rho_{i} \geq \rho$ $ \forall \, i$ \cite{Smola00Large}. Under some
regularity conditions, a large margin leads to theoretical
guarantees of good performance on new decision
instances \cite{Vapnik95TheNature}. The decision function maximizing
the margin on the training data is thus obtained by solving
\begin{equation}
\maximize_{\bw, b} \rho = \maximize_{\bw, b} \min_i
\rho_f(\bx_{i},y_i) \, . \label{eq:maxmargin}
\end{equation}
or, equivalently,
\begin{equation}
\begin{aligned}
\maximize_{\mathbf{w}, b} & \;\; \rho\\
\mbox{subject to} & \;\; \rho_{i} \geq \rho, \;\; i \in \mathcal{I} \,.
%\label{eq:maxmargin2}
\end{aligned}
\end{equation}
However, without some constraint on the size of $\bw$, the solution
to this maximin problem does not
exist \cite{Smola00Large,Hastie01Elements}. On the other hand, even
if we fix the norm of $\bw$, a separating hyperplane guaranteeing a
positive margin $\rho_f(\bx_{i},y_i)$ on each training pattern need
not exist. This is the case, for example, if a high noise level
causes a large overlap of the classes. In this case, the hyperplane
maximizing (\ref{eq:maxmargin}) performs poorly because the
prediction mechanism is determined entirely by misclassified
examples and the theoretical results guaranteeing a good
classification accuracy on unseen patterns no longer
hold \cite{Smola00Large}. A standard approach to deal with noisy
training patterns is to allow for the possibility of examples
violating the constraint $\rho_{i} \geq \rho$ $\forall \, i$ and by
computing the margin on a subset of training examples. The exact way
by which SVMs address these problems gives rise to specific
formulations, called \emph{soft-margin} SVMs.

%optimizing a regularized risk functional which balances error on the
%training examples and sparsity of the
%solution.\cite{Wahba00,Hofmann2008} Under some regularity
%conditions, this simple principle leads to theoretical guarantees of
%good performance on new decision instances.\cite{Vapnik95TheNature}
%The measure of error on the training examples is
\subsection{Soft-Margin SVM Formulations}

In $L_1$-SVMs (see e.g. \cite{Cortes95SVMs,Scholkopf01Kernels,Hastie01Elements}), degeneracy
of problem (\ref{eq:maxmargin}) is addressed by scaling the
constraints $\rho_{i} \geq \rho$ as $\tfrac{\rho_{i}}{\|\bw\|} \geq
\rho$ and by adding the constraint  $\|\bw\| = \tfrac{1}{\rho}$, so
that the problem now takes the form of the quadratic programming
problem
\begin{equation}\label{eq:PRIMAL-L1}
\begin{aligned}
\minimize_{\mathbf{w}, b} &\;\; \tfrac{1}{2} \|\bw\|^{2}\\
\mbox{subject to} &\;\; \rho_f(\bx_{i},y_i) \geq 1, \;\; i\in \mathcal{I} \,.
\end{aligned}
\end{equation}
Noisy training examples are handled by incorporating slack
variables $\xi_i \geq 0$ to the constraints in (\ref{eq:PRIMAL-L1})
and by penalizing them in the objective function:
\begin{equation}
\begin{aligned}
\minimize_{\mathbf{w}, b,\bm{\xi}} &\;\; \tfrac{1}{2} \|\bw\|^{2} + C \sum_{i \in \mathcal{I}} \xi_i\\
\mbox{subject to} &\;\; \rho_f(\bx_{i},y_i) \geq 1 - \xi_i, \;\;
i \in \mathcal{I}\,.
\end{aligned}
\end{equation}
This leads to the so called \emph{soft-margin} $L_1$-SVM. In this
formulation, the parameter $C$ controls the trade-off between margin
maximization and margin constraints violation.

Several other reformulations of problem
(\ref{eq:maxmargin}) can be found in literature. In particular, in
some formulations the two--norm of $\bm{\xi}$ is penalized instead
of the one--norm. In this article, we are particularly interested in
the {\it soft margin} $L_2$-SVM proposed by Lee and Mangasarian in
\cite{LM}. In this formulation, the margin constraints
$\rho_{i} \geq \rho$ in (\ref{eq:maxmargin}) are preserved, the
margin variable $\rho$ is explicitly incorporated in the objective
function and degeneracy is addressed by penalizing the squared 
norms of both $\mathbf{w}$ and $b$,

\begin{equation}\label{L2SVM}
\begin{aligned}
\minimize_{\mathbf{w}, b, \rho, \bm{\xi}} & \;\; \tfrac{1}{2}\left(\|\mathbf{w}\|^2 + b^2 + C \sum_{i \in \mathcal{I}} \xi_i^2 \right) - \rho \\
\mbox{subject to} & \;\; \rho_f(\bx_{i},y_i) \geq \rho - \xi_i, \;\;
i \in \mathcal{I} \, .
\end{aligned}
\end{equation}

In practice, $L_2$-SVMs and $L_1$-SVMs usually obtain a similar
classification accuracy in predicting unseen patterns \cite{LM,coreSVMs05tsang}.

\subsection{The Target QP}

In this paper we focus on the $L_2$-SVM model as described above. The use of this formulation is mainly motivated by efficiency: by adopting the slightly modified functional of Eqn. \ref{L2SVM}, we can exploit the framework introduced in \cite{coreSVMs05tsang} and solve the learning problem more easily, as we will explain in the next Subsection. As a drawback, the constraints of problem (\ref{L2SVM}) explicitly depend on the images $\bp_i = \phi(\bx_{i})$ of the training examples under the mapping $\phi$. In practice, to avoid the explicit
computation of the mapping, it is convenient to derive the Wolfe
dual of the problem by incorporating multipliers $\alpha_i \geq 0,
i \in \mathcal{I}$ and considering its Lagrangian
\begin{equation}
L(\bw,b,\bm{\xi},\bm{\alpha}) = \tfrac{1}{2}\left(\|\mathbf{w}\|^2 +
b^2 + C \sum_{i\in \mathcal{I}} \xi_i^2\right) - \rho - \sum_{i \in \mathcal{I}} \alpha_i
\left(\rho_f(\bx_{i},y_i) - \rho + \xi_i \right)\,.
\label{eq:lag-L2SVM}
\end{equation}
From the Karush-Kuhn-Tucker conditions for the optimality of (\ref{L2SVM}) with
respect to the primal variables we have
(see \cite{Cortes95SVMs,Scholkopf01Kernels,Hastie01Elements}):
\begin{equation}
\begin{aligned}
\frac{\partial L}{\partial \bw} = 0 & \Leftrightarrow \bw = \sum_{i\in \mathcal{I}}
\alpha_i y_i \phi(\bx_i) \label{eq:kkt-conditions-l2svm}\\
\frac{\partial L}{\partial b} = 0 & \Leftrightarrow b = \sum_{i\in \mathcal{I}}
\alpha_i y_i \\
\frac{\partial L}{\partial \xi_i} = 0 & \Leftrightarrow \xi_i =
\frac{\alpha_i}{C}, \;\; i\in \mathcal{I} \\
\frac{\partial L}{\partial \rho} = 0 & \Leftrightarrow \sum_{i\in \mathcal{I}}
\alpha_i = 1 \, .
\end{aligned}
\end{equation}
Plugging into the Lagrangian, we have
\begin{equation}
L(\bw,b,\bm{\xi},\bm{\alpha}) = -\frac{1}{2} \sum_{i,j \in \mathcal{I}} \alpha_i \alpha_{j} y_i y_j 
\left(\bp_i^{T}\bp_j + 1 \right) -\frac{1}{2}
\sum_{i \in \mathcal{I}}\frac{\alpha_i^{2}}{C} \, .
\end{equation}
By definition of Wolfe dual (see \cite{Scholkopf01Kernels}), it immediately follows that (\ref{L2SVM}) is equivalent to the following QP:
\begin{equation}\label{lindualL2SVM-feat}
\begin{aligned}
\maximize_{\bm{\alpha}} &\;\; -\sum_{i,j \in \mathcal{I}} \alpha_i \alpha_j \left(y_i y_j \mathbf{p}_i^{T}\mathbf{p}_j + y_i y_j + \frac{\delta_{ij}}{C} \right) \\
\mbox{subject to}& \;\; \sum_{i \in \mathcal{I}} \alpha_{i} = 1, \ \, \alpha_{i} \geq 0,  \;\; i \in \mathcal{I} \, ,
\end{aligned}
\end{equation}
where $\delta_{ij}$ is equal to 1 if $i=j$, and 0 otherwise. In
contrast to (\ref{L2SVM}), the problem above depends on the training
examples images $\bp_i = \phi(\bx_{i})$
only through the dot products $\mathbf{p}_i^{T}\mathbf{p}_j$. By
using the kernel function we can hence obtain a problem defined
entirely on the original data
\begin{equation}\label{dualL2SVM}
\begin{aligned}
\maximize_{\bm{\alpha}}  &\;\; \Theta(\bm{\alpha}) := - \sum_{i,j \in \mathcal{I}} \alpha_i \alpha_j \left(y_i y_j k(\bx_i,\bx_j) + y_i y_j + \frac{\delta_{ij}}{C} \right) \\
\mbox{subject to}& \;\; \sum_{i \in \mathcal{I}} \alpha_{i} = 1, \ \, \alpha_{i} \geq 0,  \;\; i \in \mathcal{I} \, .
\end{aligned}
\end{equation}
From equations (\ref{eq:kkt-conditions-l2svm}), we can also write the
decision function (\ref{eq:decfunc}) in terms of the original
training examples as $f(\mathbf{x}) = \mbox{sgn}\left(h(\bx)\right)$, where 
\begin{equation}\label{eq:decfunc-kernel} 
h(\mathbf{x}) = \mathbf{w}^{T}\phi(\mathbf{x}) + b = \sum_{i \in \mathcal{I}}
\alpha_i y_i \left(k(\bx_i,\bx) + 1\right) \, .
\end{equation}
Note that the decision function above depends only on the subset of training examples for which $\alpha_i \neq 0$. These examples are
usually called \emph{the support vectors} of the model \cite{Scholkopf01Kernels}. The set of support vectors is often considerably smaller than the original training set.

\subsection{Computing SVMs as Minimal Enclosing Balls (MEBs)}

Now we explain why the $L_2$-SVM formulation introduced in the previous paragraphs can lead to efficient algorithms to extract SVM classifiers from data.
As pointed out first in \cite{coreSVMs05tsang} and then generalized
in \cite{coreSVMs-generalized06tsang}, the $L_2$-SVM can be
equivalently formulated as a MEB problem in a certain feature space,
that is, as the computation of the ball of smallest radius
containing the image of the dataset under a mapping into a
dot product space $\mathcal{Z}$.

Consider the image of the training set $S$ under a mapping
$\varphi$, that is, $\varphi(S) =\{\bz_i = \varphi(\bxi{i}) : i  \in \mathcal{I} \}$. Suppose now that there exists a kernel function
$\tilde{k}$ such that $\tilde{k}(\bxi{i},\bxi{j}) =
\varphi(\bxi{i})^{T} \varphi(\bxi{j})$ $\forall \, i,j \in \mathcal{I}$. Denote
the closed ball of center $\bc \in \mathcal{Z}$ and radius $r \in
\mathbb{R}^{+}$ as $\cB(\bc,r)$. The MEB $\cB(\mathbf{c}^{\ast},r^{\ast})$
of $\varphi(S)$ can be defined as the solution of the following
optimization problem
\begin{equation}
\begin{aligned}
\minimize_{r^{2},\mathbf{c}}& \; \ r^{2} \label{eq:ballProblemPrimal}\\
\mbox{subject to}& \; \ \|\bz_i - \mathbf{c}\|^{2} \leq r^{2}, \;\; i \in \mathcal{I} \, .
\end{aligned}
\end{equation}
By using the kernel function $\tilde{k}$ to implement
dot products in $\mathcal{Z}$, the following Wolfe dual of the MEB problem
is obtained (see \cite{yildirim08}):
\begin{equation}
\begin{aligned}
\maximize_{\bm{\alpha}}&\; \ \Phi(\bm{\alpha}) := \sum_{i \in \mathcal{I}}\alpha_{i}\tilde{k}(\bxi{i},\bxi{i}) - \sum_{i,j \in \mathcal{I}} \alpha_{i}\alpha_{j}\tilde{k}(\bxi{i},\bxi{j}) \label{eq:ballProblemDual}\\
\mbox{subject to}& \; \ \sum_{i \in \mathcal{I}} \alpha_{i} = 1, \ \, \alpha_{i} \geq 0,  \;\; i \in \mathcal{I} \, .
\end{aligned}
\end{equation}
\indent If we denote by $\bm{\alpha^{\ast}}$ the solution of
(\ref{eq:ballProblemDual}), formulas for the center $\mathbf{c}^{\ast}$ and the
squared radius $r^{\ast2}$ of MEB$(\varphi(S))$ follow from
strong duality:

\begin{equation}
\mathbf{c}^{\ast} = \sum_{i \in \mathcal{I}}\alpha_{i}^{\ast}\mathbf{z}_i \, , \;\; r^{\ast2} = \Phi(\bm{\alpha^{\ast}}) = \sum_{i,j \in \mathcal{I}}\alpha_{i}^{\ast}\alpha_{j}^{\ast}\varphi(\bxi{i})^{T}\varphi(\bxi{j}) \, . \label{eq:radius-MEB}
\end{equation}

Note that the MEB depends only on the subset of points $\mathcal{C}$ for which $\alpha^{\ast}_i \neq 0$. It can be shown that computing the MEB on $\mathcal{C} \subset \varphi(S)$ is equivalent to computing the MEB on the entire dataset $\varphi(S)$. This set is frequently called a \emph{coreset} of $\varphi(S)$, a concept we are going to explore further in the next sections.

We immediately notice a deep similarity between problems (\ref{dualL2SVM}) and (\ref{eq:ballProblemDual}), the only difference being the presence of a linear
term in the objective function of the latter. This linear term can
be neglected under mild conditions on the kernel function
$\tilde{k}$. Suppose $\tilde{k}$ fulfills the following
normalization condition:

\begin{equation}
\tilde{k}(\bxi{i},\bxi{i}) =  \Delta^{2} = \mbox{constant} \,
. \label{eq:hypothesis-kernel}
\end{equation}

Since $\sum_{i\in \mathcal{I}} \alpha_i = 1$, the linear term $\sum_{i \in
\mathcal{I}}\alpha_{i}\tilde{k}(\bxi{i},\bxi{i})$ in
(\ref{eq:ballProblemDual}) becomes a constant and can be ignored when optimizing for $\bm{\alpha}$. Equivalence between the
solutions of problems (\ref{dualL2SVM}) and
(\ref{eq:ballProblemDual}) follows if we set $\tilde{k}$ to

\begin{equation}
\tilde{k} (\bxi{i},\bxi{j}) = y_i y_j \left(k(\bxi{i},\bxi{j}) + 1 \right ) + \frac{\delta_{ij}}{C} \, ,
\label{eq:kernel-L2-SVM}
\end{equation}
where $k$ is the kernel function used within the SVM
classifier. Therefore, computing an SVM for a set of labelled data $S=
\{\mathbf{x}_{i}: i \in \mathcal{I} \}$ is equivalent to computing the
MEB of the set of feature points $\varphi(S) =\{\bz_i =
\varphi(\bxi{i}) : i \in \mathcal{I}\}$, where the mapping $\varphi$
satisfies the condition $\tilde{k}(\bxi{i},\bxi{j}) =
\varphi(\bx_i)^{T}\varphi(\bx_j)$. A possible implementation of such
a mapping is $\varphi(\bx_i) = \left(y_i
\phi(\bx_i),y_i,\tfrac{1}{\sqrt{c}} \bm{e}_i\right)$, where
$\phi(\bx_i)$ is in turn the mapping associated with the original
Mercer kernel $k$ used by the SVM.

Note that the previous equivalence between the MEB and the
SVM problems holds if and only if the kernel $\tilde{k}$
fulfills assumption (\ref{eq:hypothesis-kernel}). If, for
example, the SVM classifier implements the well-known $d$-th order
polynomial kernel $k(\bx_i,\bx_j) = \left(\bx_i^{T}\bx_j +
1\right)^{d}$, we have that $\tilde{k}(\bxi{i},\bxi{i})$ is no
longer a constant, and thus the MEB equivalence no longer holds.
Complex constructions are required to extend the MEB optimization
framework to SVMs using different kernel
functions \cite{coreSVMs-generalized06tsang}.

\section{B\u{a}doiu-Clarkson Algorithm and Core Vector Machines}

Problem (\ref{eq:ballProblemDual}) is in general a large and dense
QP. Obtaining a numerical solution when $m$ is large is very
expensive, no matter which kind of numerical method one decides to
employ. Taking into account that in practice we can only approximate
the solution within a given tolerance, it is convenient to modify a
priori our objective: instead of MEB$(\varphi(S))$, we can try to
compute an approximate MEB in the sense specified by the following
definition.
\begin{definition}
Let MEB$(\varphi(S))$=$\mathcal{B}(\mathbf{c}^{\ast},r^{\ast})$ and
$\epsilon > 0$ be a given tolerance. Then, a $(1+\epsilon)$--MEB of
$\varphi(S)$ is a ball $\mathcal{B}(\mathbf{c},r)$ such that
\begin{equation}
r \leq r^{\ast} \;\; \mbox{and} \;\; \varphi(S) \subset \mathcal{B}(\mathbf{c},(1+\epsilon)r) \,.
\end{equation}
A set $\mathcal{C}_S \subset \varphi(S)$ is an $\epsilon$--\emph{coreset} of $\varphi(S)$ if MEB$(\mathcal{C}_S)$ is a $(1+\epsilon)$--MEB of $\varphi(S)$.
\end{definition}

In \cite{BadoiuClarkson03smaller-coresets} and
\cite{yildirim08}, algorithms to compute $(1+\epsilon)$--MEBs
that scale independently of the dimension of $\cZ$ and the
cardinality of $S$ have been provided. In particular, the
B\u{a}doiu-Clarkson (BC) algorithm described in
\cite{BadoiuClarkson03smaller-coresets} is able to provide an
$\epsilon$--coreset $\mathcal{C}_S$ of $\varphi(S)$ in no more than
$O(1/\epsilon)$ iterations. We denote with $\mathcal{C}_k$ the coreset approximation obtained at the $k$-th iteration and its MEB as
$B_k = \cB(\bc_k,r_k)$. Starting from a given $\mathcal{C}_0$, at
each iteration $\mathcal{C}_{k+1}$ is defined as the union of
$\mathcal{C}_k$ and the point of $\varphi(S)$ furthest from $\mathbf{c}_{k}$.
The algorithm then computes $B_{k+1}$ and stops if
$\cB(\mathbf{c}_{k+1},(1+\epsilon)r_{k+1})$ contains $\varphi(S)$.

Exploiting these ideas, Tsang and colleagues introduced in
\cite{coreSVMs05tsang} the CVM (Core Vector Machine) for training
SVMs supporting a reduction to a MEB problem. The CVM is described in
Algorithm \ref{alg:cvms}, where each $\mathcal{C}_k$ is identified
by the index set $\mathcal{I}_k \subset \mathcal{I}$. The elements
included in $\mathcal{C}_k$ are called the \emph{core vectors}.
Their role is exactly analogous to that of support vectors in a
classical SVM model.

The expression for the radius $r_{k}$ follows easily from
(\ref{eq:radius-MEB}). Moreover, it is easy to show (see
\cite{coreSVMs05tsang}) that step \ref{cvms:istar} exactly looks
for the point $\bxi{i^{\ast}}$ whose image $\varphi(\bxi{i^{\ast}})$
is the furthest from $\mathbf{c}_{k}$. In fact, by using the
expressions $\mathbf{c}_k = \sum_{j \in \mathcal{I}_k} \alpha_{k,j}
\mathbf{z}_j$ and $\tilde k(\mathbf{x}_i,\mathbf{x}_i) = \Delta^2$
$\forall \, i \in \mathcal{I}$, we obtain:

\begin{equation}\label{distance}
\begin{aligned}
\|  \mathbf{z}_i - \mathbf{c}_k\|^2 &=  \Delta^2 + \sum_{j,l \in \mathcal{I}_k} \alpha_{k,j} \alpha_{k,l} \tilde k(\mathbf{x}_j,\mathbf{x}_l)-2 \sum_{j \in \mathcal{I}_k} \alpha_{k,j}
 \tilde k(\mathbf{x}_j,\mathbf{x}_i) \\
 &= \Delta^2 + R_k - 2 \sum_{j \in \mathcal{I}_k} \alpha_{k,j}
 \tilde k(\mathbf{x}_j,\mathbf{x}_i) \,.
 \end{aligned}
\end{equation}

Note how this computation can be performed, by means of kernel
evaluations, in spite of the lack of an explicit representation of
$\mathbf{c}_k$ and $\mathbf{z}_i$. Once $i^{\ast}$ has been found,
it is included in the index set. Finally, the reduced QP
corresponding to the MEB of the new approximate coreset is solved.

Algorithm \ref{alg:cvms} has two main sources of computational
overhead: the computation of the furthest point from $\mathbf{c}_k$,
which is linear in $m$, and the solution of the optimization
subproblem in step \ref{cvms:nestedQP}. The complexity of the former
step can be made constant and independent of $m$ by suitable
sampling techniques (see \cite{coreSVMs05tsang}), an issue to which we will
return later. As regards the optimization step, CVMs adopt a SMO
method, where only
two variables are selected for optimization at each iteration \cite{smo-second-order05fan,platt99smo-seminal}. It is
known that the cost of each SMO iteration is not too high, but the
method can require a large number of iterations in order to satisfy
reasonable stopping criteria \cite{platt99smo-seminal}.

\begin{algorithm}
\caption{BC Algorithm for MEB-SVMs: the Core Vector Machine}
\label{alg:cvms}
\begin{algorithmic}[1]
\Require $S$, $\epsilon$.%\\
\State \textbf{initialization:} compute $\mathcal{I}_0$ and
$\bm{\alpha}_0$;
\State $\Delta^{2} \longleftarrow \tilde{k}(\bxi{1},\bxi{1})$;%\; \\
\State $R_0 \longleftarrow \sum_{i,j \in \mathcal{I}_{0}} \alpha_{0,i}\alpha_{0,j} \tilde{k}(\bxi{i},\bxi{j})$;% \\
\State $r_0^2 \longleftarrow  \Delta^2 - R_0$;%\;\\
\State $i^{\ast} \longleftarrow \argmax_{i \in \mathcal{I}} \gamma^2(\bm{\alpha}_0;i) := \Delta^{2} + R_0 - 2 \sum_{j \in \mathcal{I}_{0}} \alpha_{0,j} \tilde{k}(\bxi{j},\bxi{i})$;%\;\\
%5. $\gamma_0^2 \longleftarrow \Delta^{2} + R_0 - 2 \sum_{j \in \mathcal{I}_{0}} \alpha_{0,j} \tilde{k}(\bxi{j},\bxi{i^{\ast}})$; \\
\State $k \longleftarrow 0$;%\; \\
\While{$\gamma^{2}(\bm{\alpha}_k;i^{\ast}) > (1+\epsilon)^2 r_k^{2}$}%
\State $k \longleftarrow k + 1$;%\;\\
\State $\mathcal{I}_{k} \longleftarrow \mathcal{I}_{k-1} \cup \{i^{\ast} \}$;%\;\\
\State Find $\label{cvms:nestedQP} \bm{\alpha}_{k}$ by solving the reduced QP problem% \\
\begin{equation}
\begin{aligned}
\minimize_{\bm{\alpha} \in \mathbb{R}^m} \;\;& R(\bm{\alpha}):=
\sum_{i,j \in \mathcal{I}_k} \alpha_{i} \alpha_{j} \tilde
k(\mathbf{x}_i,\mathbf{x}_j)
\\ \mbox{subject to} \;\; &\sum_{i \in \mathcal{I}_{k}} \alpha_{i} = 1, \ \alpha_{i} \geq 0, \;\; i \in \mathcal{I}_{k} \,;%\;
\end{aligned}
\end{equation}
\State $R_k \longleftarrow R(\bm{\alpha}_k)$;% \\
\State $r_{k}^{2} \longleftarrow \Delta^{2} - R_k$;%\; \\
\State $\label{cvms:istar} i^{\ast} \longleftarrow \argmax_{i \in \mathcal{I}} \gamma^2(\bm{\alpha}_k;i) := \Delta^{2} + R_k - 2 \sum_{j \in \mathcal{I}_{k}} \alpha_{k,j} \tilde{k}(\bxi{j},\bxi{i})$;%\; \\
%7.7. $ \gamma_k^2 \longleftarrow \Delta^{2} + R_k - 2 \sum_{j \in \mathcal{I}_{k}} \alpha_{k,j} \tilde{k}(\bxi{j},\bxi{i^{\ast}})$; \\
\EndWhile%\\
\State \textbf{return} $\mathcal{I}_{S}=\mathcal{I}_{k}$, $\bm{\alpha} = \bm{\alpha}_{k}$.%\\
\end{algorithmic}
\end{algorithm}

As regards the initialization, that is, the computation of
$\mathcal{C}_0$ and $\bm{\alpha}_0$, a simple choice is suggested in
\cite{Kumar03coresets}, which consists in choosing
$\mathcal{C}_0=\{\mathbf{z}_a, \mathbf{z}_b\}$, where $\mathbf{z}_a$
is an arbitrary point in $\varphi(S)$ and $\mathbf{z}_b$ is the farthest
point from $\mathbf{z}_a$. Obviously, in this case the center and
radius of $B_0$ are $\mathbf{c}_0=0.5(\mathbf{z}_a+\mathbf{z}_b)$ and
$r_0=0.5\|\mathbf{z}_a-\mathbf{z}_b\|$, respectively. That is, we
initialize $\mathcal{I}_0 = \{a,b\}$,
$\alpha_{0,a}=\alpha_{0,b}=0.5$ and $\alpha_{0,i}=0$ for $i \notin
\mathcal{I}_0$. A more efficient strategy, implemented for example
in the code LIBCVM \cite{LibCVM09}, is the following. The procedure
consists in determining the MEB of a subset $P=\{\mathbf{z}_i, i \in
\mathcal{I}_P\}$ of $p$ training points, where the set of indices
$\mathcal{I}_P$ is randomly chosen and $p$ is small. This MEB is
approximated by running a SMO solver. In practice, $p \simeq 20$ is
suggested to be enough, but one can also try larger initial guesses,
as long as SMO can rapidly compute the initial MEB. $\mathcal{C}_0$
is then defined as the set of points $\mathbf{x}_i \in P$ gaining a
strictly positive dual weight in the process, and $\mathcal{I}_0$ as
the set of the corresponding indices.

\section{Frank-Wolfe Methods for the MEB-SVM Problem}

\subsection{Overview of the Frank-Wolfe Algorithm}

The \textit{Frank-Wolfe algorithm} (FW), originally presented in \cite{wolfe1954}, is designed to solve optimization problems of the form
\begin{equation} \label{generalpb}
\maximize_{\bm{\alpha} \in \Sigma} f(\bm{\alpha}) \,,
\end{equation}
where $f \in C^1(\mathbb{R}^m)$ is a concave function, and $\Sigma \neq \emptyset$ a bounded convex polyhedron.

In the case of the MEB dual problem, the objective function is
quadratic and $\Sigma$ coincides with the unit simplex. Given the
current iterate $\bm{\alpha}_k \in \Sigma$, a standard Frank-Wolfe
iteration consists in the following steps:
\begin{enumerate}
\item Find a point $\mathbf{u}_k \in \Sigma$ maximizing the local linear approximation $\psi_k(\mathbf{u}):= f(\bm{\alpha}_k) + (\mathbf{u} - \bm{\alpha}_k)^T \nabla f(\bm{\alpha}_k)$, and define $\mathbf{d}_k^{FW} = \mathbf{u}_k - \bm{\alpha}_k$. \label{linapprox}
\item Perform a line-search $\lambda_k =  \argmax_{\lambda \in [0,1]}  f(\bm{\alpha}_k + \lambda
\mathbf{d}_k^{FW})$. \label{linesearch}
\item Update the iterate by
\begin{equation}\label{alphanew}
\bm{\alpha}_{k+1}=  \bm{\alpha}_k + \lambda_k \mathbf{d}_k^{FW} = (1-\lambda_k)\bm{\alpha}_k + \lambda_k \mathbf{u}_k \,.
\end{equation}
\end{enumerate}
The algorithm is usually stopped when the objective function is sufficiently close to its optimal value, according to a suitable proximity measure \cite{GuelatMarcotte}.

Since $\psi_k(\mathbf{u})$ is a linear function and $\Sigma$ is a
bounded polyhedron, the search directions $\mathbf{d}_k^{FW}$ are
always directed towards an extreme point of $\Sigma$. That is,
$\mathbf{u}_k$ is a vertex of the feasible set. The constraint
$\lambda_k \in [0,1]$ ensures feasibility at each iteration. It is
easy to show that in the case of the MEB problem $\mathbf{u}_k =
\mathbf{e}_{i^{\ast}}$, where $\mathbf{e}_i$ denotes the $i$-th
vector of the canonical basis, and $i^{\ast}$ is the index
corresponding to the largest component of $\nabla
f(\bm{\alpha}_k)$ \cite{yildirim08}. The updating step therefore
assumes the form
\begin{equation}\label{FWupdate}
\bm{\alpha}_{k+1} = (1 - \lambda_k)\bm{\alpha}_k + \lambda_k \mathbf{e}_{i^{\ast}} \,.
\end{equation}

It can be proved that the above procedure converges globally \cite{GuelatMarcotte}. As a drawback, however, it often exhibits a tendency to stagnate near a solution. Intuitively, suppose that solutions $\bm{\alpha}^{\ast}$ of (\ref{generalpb}) lie on the boundary of $\Sigma$ (this is often true in practice, and holds in particular for the MEB problem). In this case, as $\bm{\alpha}_k$ gets close to a solution $\bm{\alpha}^{\ast}$, the directions $\mathbf{d}_k^{FW}$ become more and more orthogonal to $\nabla f(\bm{\alpha}_k)$. As a consequence, $\bm{\alpha}_k$ possibly never reaches the face of $\Sigma$ containing $\bm{\alpha}^{\ast}$, resulting in a sublinear convergence rate \cite{GuelatMarcotte}.

\subsection{The Modified Frank-Wolfe Algorithm}

We now describe an improvement over the general Frank-Wolfe procedure, which was first proposed in \cite{wolfe1970} and later detailed in \cite{GuelatMarcotte}. This improvement can be quantified in terms of the rate of convergence of the algorithm and thus of the number of iterations in which it can be expected to fulfill the stopping conditions. 

In practice, the tendency of FW to stagnate near a solution can lead to later iterations wasting computational resources while making minimal progress towards the optimal function value. It would thus be desirable to obtain a stronger result on the convergence rate, which guarantees that the speed of the algorithm does not deteriorate when approaching a solution. This paragraph describes a technique geared precisely towards this aim.

Essentially, the previous algorithm is enhanced by introducing alternative search directions known as \textit{away steps}. The basic idea is that, instead of moving towards the vertex of $\Sigma$ maximizing a linear approximation $\psi_k$ of $f$ in $\bm{\alpha}_k$, we can move away from the vertex minimizing $\psi_k$. At each iteration, a choice between these two options is made by choosing the best ascent direction. The whole procedure, known as \textit{Modified Frank-Wolfe algorithm} (MFW), can be sketched as follows:
\begin{enumerate}
\item Find $\mathbf{u}_k \in \Sigma$ and define $\mathbf{d}_k^{FW}$ as in the standard FW algorithm.
\item Find $\mathbf{v}_k \in \Sigma$ by minimizing $\psi_k(\mathbf{v})$, s.t. $v_j=0$ if $\alpha_{k,j} = 0$. Define $\mathbf{d}_k^{A} = \bm{\alpha}_k - \mathbf{v}_k$.\label{minstep}
\item If $\nabla f(\bm{\alpha}_k)^T \mathbf{d}_k^{FW} \geq \nabla f(\bm{\alpha}_k)^T \mathbf{d}_k^{A}$, then $\mathbf{d}_k = \mathbf{d}_k^{FW}$, else $\mathbf{d}_k = \mathbf{d}_k^{A}$.
\item Perform a line-search $\lambda_k =  \argmax_{\lambda \in [0,\bar \lambda]}  f(\bm{\alpha}_k + \lambda \mathbf{d}_k)$, where $\bar \lambda = 1$ if $\mathbf{d}_k = \mathbf{d}_k^{FW}$ and $\bar \lambda = \max_{\lambda \geq 0} \left \{\lambda \,|\, \bm{\alpha}_k + \lambda \mathbf{d}_k^{A} \in \Sigma \right \}$.
\item Update the iterate by
\begin{equation}\label{alphanew}
\bm{\alpha}_{k+1}=  \bm{\alpha}_k + \lambda_k \mathbf{d}_k =  \left \{
\begin{aligned}
 (1-\lambda_k)\bm{\alpha}_k + \lambda_k \mathbf{u}_k & \qquad \text{if} &  \mathbf{d}_k&= \mathbf{d}_k^{FW} \\
 (1+\lambda_k)\bm{\alpha}_k - \lambda_k \mathbf{v}_k & \qquad \text{if} &  \mathbf{d}_k&= \mathbf{d}_k^{A} \,.
\end{aligned}
\right.
\end{equation}
\end{enumerate}
It is easy to show that both $\mathbf{d}_k^{FW}$ and $\mathbf{d}_k^{A}$ are feasible ascent directions, unless $\bm{\alpha}_k$ is already a stationary point.

In the case of the MEB problem, step \ref{minstep} corresponds to finding the basis vector $\mathbf{e}_{j^{\ast}}$ corresponding to the smaller component of $\nabla f (\bm{\alpha}_k)$ \cite{yildirim08}. Note that a face of $\Sigma$ of lower dimensionality is reached whenever an away step with maximal stepsize $\bar \lambda$ is performed. Imposing the constraint in step \ref{minstep} is tantamount to ruling out away steps with zero stepsize. That is, an away step from $\mathbf{e}_j$ cannot be taken if $\alpha_{k,j}$ is already zero.

%%%BEGIN MODIFICHE%%%
In \cite{GuelatMarcotte} linear convergence of $f(\bm{\alpha}_k)$ to $f(\bm{\alpha}^{\ast})$ was proved, assuming Lipschitz continuity of $\nabla f$, strong concavity of $f$, and strict complementarity at the solution. In \cite{yildirim08}, a proof of the same result was provided for the MEB problem, under weaker assumptions. 
It is important to note that such assumptions are in particular satisfied for the MEB formulation of the $L_2$-SVM, and that as such the aforementioned linear convergence property holds for all problems considered in this paper.
%%%END MODIFICHE%%%
In particular, uniqueness of the solution, which is implied if we ask for strong (or just strict) concavity, is not required. The gist is essentially that, in a small neighborhood of a solution $\bm{\alpha}^{\ast}$, MFW is forced to perform away steps until the face of $\Sigma$ containing $\bm{\alpha}^{\ast}$ is reached, which happens after a finite number of iterations. Starting from that point, the algorithm behaves as an unconstrained optimization method, and it can be proved that $f(\bm{\alpha}_k)$ converges to $f(\bm{\alpha}^{\ast})$ linearly \cite{GuelatMarcotte}.

\subsection{The FW and MFW Algorithms for MEB-SVMs}

If FW method is applied to the MEB dual problem, the structure of
the objective function $\Phi(\bm{\alpha})$ can be exploited in order
to obtain explicit formulas for steps \ref{linapprox} and
\ref{linesearch} of the generic procedure. Indeed, the components
of $\nabla\Phi(\bm{\alpha}_k)$ are given by
\begin{equation}
\nabla \Phi(\bm{\alpha}_k)_i = \|\mathbf{z}_i\|^2 - 2 \sum_{j \in \mathcal{I}} \alpha_{k,j} \mathbf{z}_i^T \mathbf{z}_j =
 \|\mathbf{z}_i\|^2 - 2 \mathbf{z}_i^T\mathbf{c}_k \,,
\end{equation}
where
\begin{equation}\label{ck}
\mathbf{c}_k=\sum_{j \in \mathcal{I}} \alpha_{k,j} \mathbf{z}_j \,,
\end{equation}
and therefore, since $\| \mathbf{c}_k \|^2$ does not depend on $i$,
\begin{equation}
i^{\ast} = \argmax_{i \in \mathcal{I}}\nabla \Phi(\bm{\alpha}_k)_i=
\argmax_{i \in \mathcal{I}}\| \mathbf{z}_i - \mathbf{c}_{k}\|^{2} .
\end{equation}
In practice, step \ref{linapprox} selects the index of the input
point maximizing the distance from $\mathbf{c}_k$, exactly as done
in the CVM procedure. Computation of distances can be carried out as in CVMs,
using (\ref{distance}). As regards step \ref{linesearch}, it can be
shown (see \cite{clarkson08coresets,yildirim08}) that
\begin{equation} \label{lambdaopt}
\lambda_k = \frac{1}{2}\left(1 - \frac{r_k^2}{\| \mathbf{z}_{i^{\ast}} - \mathbf{c}_k\|^2}\right),
\end{equation}
where
\begin{equation}\label{rhok}
r^2_k=\Phi(\bm{\alpha}_k) \,.
\end{equation}
By comparing (\ref{ck}) and (\ref{rhok}) with (\ref{eq:radius-MEB}), we argue that, as in the BC algorithm, a ball $B_k=\mathcal{B}(\mathbf{c}_k, r_k)$ is identified at each iteration.

The whole procedure is sketched in Algorithm \ref{alg:fvms}, where
at each iteration we associate to $\bm{\alpha}_k$  the index set
$\mathcal{I}_k= \{ i \in \mathcal{I} : \alpha_{k,i} >0\}$.

\begin{algorithm}[ht]
\caption{Frank-Wolfe Algorithm for MEB-SVMs} \label{alg:fvms}
\begin{algorithmic}[1]
\Require $S$, $\epsilon$.%
\State \textbf{initialization:} compute $\mathcal{I}_0$ and
$\bm{\alpha}_0$;
\State $\Delta^{2} \longleftarrow \tilde{k}(\bxi{1},\bxi{1})$;%
\State $R_0 \longleftarrow \sum_{i,j \in \mathcal{I}_{0}} \alpha_{0,i} \alpha_{0,j} \tilde{k}(\bxi{i},\bxi{j})$; %
\State $r_0^2 \longleftarrow  \Delta^2 - R_0$;%
\State $i^{\ast} \longleftarrow \argmax_{i \in \mathcal{I}} \gamma^2(\bm{\alpha}_0;i) := \Delta^{2} + R_0 - 2 \sum_{j \in \mathcal{I}_{0}} \alpha_{0,j} \tilde{k}(\bxi{j},\bxi{i})$;%
%5. $\gamma_0^2 \longleftarrow \Delta^{2} + R_0 - 2 \sum_{j \in \mathcal{I}_{0}} \alpha_{0,j} \tilde{k}(\bxi{j},\bxi{i^{\ast}})$; \\
\State $\delta_0 \longleftarrow \frac{\gamma^2(\bm{\alpha}_0;i^{\ast})}{r^2_0}-1$;%
\State $k \longleftarrow 0$;%
\While{$ \delta_k > (1+ \epsilon)^2 - 1$}
\State $\lambda_k \longleftarrow \frac{1}{2}\left(1 - \frac{r_k^2}{\gamma^2(\bm{\alpha}_k;i^{\ast})}\right)$;%
\State $k \longleftarrow k+1$;%
\State $\bm{\alpha}_k \longleftarrow (1-
\lambda_{k-1})\bm{\alpha}_{k-1} + \lambda_{k-1} \mathbf{e}_{i^{\ast}}$;%
\State $\mathcal{I}_{k} \longleftarrow \{ i \in \mathcal{I} : \alpha_{k,i} >0\}$;%
\State $R_k \longleftarrow \sum_{i,j \in \mathcal{I}_{k}} \alpha_{k,i} \alpha_{k,j} \tilde{k}(\bxi{i},\bxi{j})$;%
\State $\label{fvms:updateradius} r^2_{k} \longleftarrow r^2_{k-1} \left(1+ \frac{\delta_{k-1}^2}{4(1+\delta_{k-1})}\right)$;%
\State $i^{\ast} \longleftarrow \argmax_{i \in \mathcal{I}} \gamma^2(\bm{\alpha}_k;i) := \Delta^{2} + R_k - 2 \sum_{j \in \mathcal{I}_{k}} \alpha_{k,j} \tilde{k}(\bxi{j},\bxi{i})$;%
%9.7. $\gamma_k^2 \longleftarrow \Delta^{2} + R_k - 2 \sum_{j \in \mathcal{I}_{k}} \alpha_{k,j} \tilde{k}(\bxi{j},\bxi{i^{\ast}})$; \\
\State $\delta_k \longleftarrow \frac{\gamma^2(\bm{\alpha}_k;i^{\ast})}{r^2_k}-1$;%
\EndWhile %
\State \textbf{return}
$\mathcal{I}_S=\mathcal{I}_k$, $\bm{\alpha}=\bm{\alpha}_k$.
\end{algorithmic}
\end{algorithm}

As regards the initialization, $\bm{\alpha}_0$ and $\mathcal{I}_0$
can be defined exactly as in the CVM procedure. At subsequent
iterations, the formula to update $\mathcal{I}_k$ immediately
follows from the updating (\ref{FWupdate}) for $\bm{\alpha}_k$;
indeed, the indices of the strictly positive components of
$\bm{\alpha}_{k+1}$ are the same of $\bm{\alpha}_k$, plus $i^{\ast}$
if $\alpha_{k,i^{\ast}}=0$ (which means that $\mathbf{z}_{i^{\ast}}$
was not already included in the current coreset). The introduction
of the sequence $\{\mathcal{I}_k\}$ in Algorithm \ref{alg:fvms}
makes it evident that structure and output of Algorithm
\ref{alg:cvms} are preserved.

The updating formula used in step \ref{fvms:updateradius} appears in
\cite{yildirim08}. It is easy to see that it is equivalent
to (\ref{rhok}) and computationally more convenient.

\begin{figure}[htb]\label{mebfigure}
\centering
\scriptsize
%\definecolor{uququq}{rgb}{0.25,0.25,0.25}
%\definecolor{qqffqq}{rgb}{0,0.5,0}
%\definecolor{qqqqff}{rgb}{0,0,1}
%\definecolor{ffqqqq}{rgb}{1,0,0}
\begin{tikzpicture}[scale=0.6,line cap=round,line join=round,>=triangle 45,x=1.0cm,y=0.9cm]
%\clip(-4.81,-5.27) rectangle (10.93,6.35);
\draw [dash pattern=on 1pt off 3pt on 5pt off 4pt] (3.19,1.38) circle (4.31cm);
\draw(2.12,0.28) circle (3.28cm);
\draw [dash pattern=on 5pt off 5pt] (2.84,0.96) circle (3.98cm);
\draw [dotted] (6.45,4.25)-- (2.12,0.28);
\fill  (6.55,4.37) circle (2pt);
\draw (6.95,4.88) node {$\mathbf{z}_{i^{\ast}}$};
\draw(-1.2,6.17) node {$B_{k+1}=\,$MEB($\mathcal{C}_{k+1}$) [BC]};
\draw (2.97,4.75) node {$B_{k+1}$ [FW]};
\draw (2.97,3.03) node {MEB$(\mathcal{C}_{k})$};
\fill  (2.12,0.28) circle (2pt);
\draw (1.69,0.02) node {$\mathbf{c}_k$};
%\draw (3.15,1.9) node {$\sqrt{\delta^k}$};
\end{tikzpicture}
 \caption{Approximating balls computed by algorithms BC and FW.}
   \label{AggiorMEB}
\end{figure}
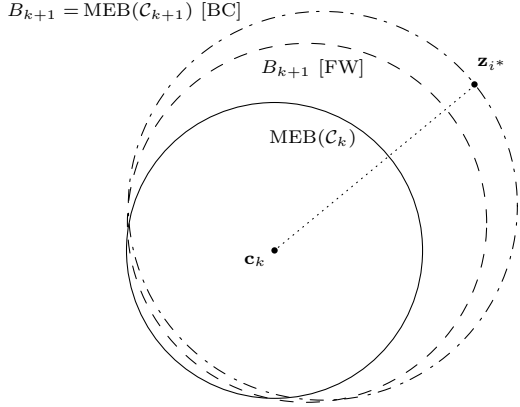

In \cite{yildirim08}, it has been proved that $\{r^2_{k}\}$ is a monotonically increasing sequence with $r^{\ast 2}$ as an upper bound. Therefore, since the same stopping criterion of the BC algorithm is used, $\mathcal{I}_S$ identifies an $\epsilon$--coreset $\mathcal{C}_S$ of $\varphi(S)$, and the last $B_k$ is a $(1+\epsilon)$--MEB of $\varphi(S)$.
However, the MEB-approximating procedure differs from that of BC in that the value of $r^2_k$ is not equal to the squared radius of MEB$(\mathcal{C}_k)$, but tends to the correct value as $\bm{\alpha}_k$ gets near the optimal solution (see Fig. \ref{AggiorMEB}).

The derivation of the MFW method applied to the MEB-SVM problem can
be written down along the same lines. Following the presentation in
\cite{yildirim08}, we describe the detailed procedure in
Algorithm \ref{alg:mfvms}.

\begin{algorithm}[h!!!]
\caption{Modified Frank-Wolfe Algorithm for MEB-SVMs}
\label{alg:mfvms}
\begin{algorithmic}[1]
\Require $S$, $\epsilon$.
\State \textbf{initialization:} compute $\mathcal{I}_0$ and
$\bm{\alpha}_0$;
\State $\Delta^{2} \longleftarrow \tilde{k}(\bxi{1},\bxi{1})$;
\State $R_0 \longleftarrow \sum_{i,j \in \mathcal{I}_{0}} \alpha_{0,i} \alpha_{0,j} \tilde{k}(\bxi{i},\bxi{j})$;%
\State $r_0^2 \longleftarrow  \Delta^2 - R_0$;%
\State $i^{\ast} \longleftarrow \argmax_{i \in \mathcal{I}} \gamma^2(\bm{\alpha}_0;i) := \Delta^{2} + R_0 - 2 \sum_{j \in \mathcal{I}_{0}} \alpha_{0,j} \tilde{k}(\bxi{j},\bxi{i})$;%\;\\
\State $j^{\ast} \longleftarrow \argmin_{j \in \mathcal{I}_0} \gamma^2(\bm{\alpha}_0;j)$;% \\
%5. $\gamma_{0+}^{2} \longleftarrow \Delta^{2} + R_0 - 2 \sum_{j \in \mathcal{I}_{0}} \alpha_{0,j} \tilde{k}(\bxi{j},\bxi{i^{\ast}})$; \\
%5. $\gamma_{0-}^{2} \longleftarrow \Delta^{2} + R_0 - 2 \sum_{j \in \mathcal{I}_{0}} \alpha_{0,j} \tilde{k}(\bxi{j},\bxi{j^{\ast}})$; \\
\State $\delta_{0+} \longleftarrow \frac{\gamma^2(\bm{\alpha}_0;i^{\ast})}{r^2_k}-1$;% \\
\State $\delta_{0-} \longleftarrow 1 - \frac{\gamma^2(\bm{\alpha}_0;j^{\ast})}{r^2_k}$;%
\State $k \longleftarrow 0$;% \\
\While{ $\delta_{k+} > (1+ \epsilon)^2 - 1$}%
\If{$\label{mfvms:choice}\delta_{k+} \geq \delta_{k-}$}
\State $\lambda_k \longleftarrow \frac{1}{2}\left(1 - \frac{r_k^2}{\gamma^2(\bm{\alpha}_k;i^{\ast})}\right)$;% \\
\State $k \longleftarrow k+1$;% \\
\State $\bm{\alpha}_k \longleftarrow (1- \lambda_{k-1})\bm{\alpha}_{k-1} + \lambda_{k-1} \mathbf{e}_{i^{\ast}}$;% \\
\State $r^2_{k} \longleftarrow r^2_{k-1} \left(1+ \frac{\delta_{k-1+}^2}{4(1+\delta_{k-1+})}\right)$; % \\
\Else%
\State $\lambda_k \longleftarrow \min \left \{ \frac{\delta_{k-}}{2(1- \delta_{k-})}, \frac{\alpha_{k,{j^{\ast}}}}{1-\alpha_{k,{j^{\ast}}}} \right \}$;% \\
\State $k \longleftarrow k+1$;% \\
\State $\label{mfvms:boundlambda} \bm{\alpha}_k \longleftarrow (1+ \lambda_{k-1})\bm{\alpha}_{k-1} - \lambda_{k-1} \mathbf{e}_{j^{\ast}}$;% \\
\State $\label{mfvms:updateradius2}r^2_{k} \longleftarrow (1 + \lambda_{k-1})r^2_{k-1} - \lambda_{k-1}(1+\lambda_{k-1})(\delta_{k-1-} - 1)r^2_{k-1}$;% \\
\EndIf
\State $\mathcal{I}_{k} \longleftarrow \{ i \in \mathcal{I} : \alpha_{k,i} >0\}$;% \\
\State $R_k \longleftarrow \sum_{i,j \in \mathcal{I}_{k}} \alpha_{k,i} \alpha_{k,j} \tilde{k}(\bxi{i},\bxi{j})$;% \\
\State $i^{\ast} \longleftarrow \argmax_{i \in \mathcal{I}} \gamma^2(\bm{\alpha}_k;i) := \Delta^{2} + R_k - 2 \sum_{j \in \mathcal{I}_{k}} \alpha_{k,j} \tilde{k}(\bxi{j},\bxi{i})$;%\;\\
\State $j^{\ast} \longleftarrow \argmin_{j \in \mathcal{I}_k}  \gamma^2(\bm{\alpha}_k;j)$;%\\
%9.6. $\gamma_{k+}^2 \longleftarrow \Delta^{2} + R_k - 2 \sum_{j \in \mathcal{I}_{k}} \alpha_{k,j} \tilde{k}(\bxi{j},\bxi{i^{\ast}})$; \\
%9.6. $\gamma_{k-}^2 \longleftarrow \Delta^{2} + R_k - 2 \sum_{j \in \mathcal{I}_{k}} \alpha_{k,j} \tilde{k}(\bxi{j},\bxi{j^{\ast}})$; \\
\State $\delta_{k+} \longleftarrow \frac{\gamma^2(\bm{\alpha}_k; i^{\ast})}{r^2_k}-1$;% \\
\State $\delta_{k-} \longleftarrow 1 - \frac{\gamma^2(\bm{\alpha}_k; j^{\ast})}{r^2_k}$;% \\
\EndWhile
\State \textbf{return} $\mathcal{I}_S=\mathcal{I}_k$, $\bm{\alpha}=\bm{\alpha}_k$.%
\end{algorithmic}
\end{algorithm}

By now, it should be apparent that $j^{\ast}$ is the index
identifying the point furthest from $\mathbf{c}_k$, and that it
corresponds to the smallest component of $\nabla
\Phi(\bm{\alpha}_k)$. That is, in Algorithm \ref{alg:mfvms} we
consider performing away steps in which the weight of the nearest
point to the current center is reduced. Of course, since the weight
of a point is not allowed to drop below zero, the search for
$j^{\ast}$ is performed on $\mathcal{I}_k$ only. Again, the optimal
stepsize can be determined in closed form \cite{yildirim08}. In
particular, it is easy to see that the expression in step
\ref{mfvms:boundlambda} corresponds to
\begin{equation}
\lambda_k =  \argmax_{\lambda \in \left [0,\frac{\alpha_{k,j^{\ast}}}{1-\alpha_{k,j^{\ast}}} \right ]} \Phi((1+\lambda)\bm{\alpha}_k - \lambda \mathbf{e}_{j^{\ast}}) \,,
\end{equation}
where the upper bound on the interval preserves dual feasibility.

This kind of step has an intuitive geometrical meaning: if we consider a solution $\bm{\alpha}^{\ast}$ of the MEB problem, it is known that nonzero components of $\bm{\alpha}^{\ast}$ correspond to points lying on the boundary of the exact MEB. Therefore, it makes sense to try to remove from the model points that lie near the center (i.e. far from the boundary of the ball).
When an away step is performed, if $\lambda_k$ is chosen as the supremum of the search interval, we get $\alpha_{k+1,j^{\ast}} = 0$ and the corresponding example is removed from the current coreset (\textit{drop step}). Moreover, it's not hard to see that step \ref{mfvms:choice} chooses to perform an away step whenever
\begin{equation}
\nabla \Phi(\bm{\alpha}_k)^T \mathbf{d}^{A}_k > \nabla \Phi(\bm{\alpha}_k)^T \mathbf{d}^{FW}_k.
\end{equation}
That is, the choice between FW and away steps is done by choosing
the best ascent direction, exactly as required by the MFW procedure.
Here $\mathbf{d}_k^{FW} = (\mathbf{e}_{i^{\ast}} - \bm{\alpha}_k)$
and $\mathbf{d}_k^A = (\bm{\alpha}_k - \mathbf{e}_{j^{\ast}})$
denote the search directions of FW and away steps, respectively.
Finally, step \ref{mfvms:updateradius2} shows that, just as with
standard FW steps, after performing an away step we can use an
analytical formula to update $r^2_k$. This expression follows easily
by writing the objective function $\Phi(\bm{\alpha})$ for
$\bm{\alpha} = (1+\lambda)\bm{\alpha}_k - \lambda
\mathbf{e}_{j^{\ast}}$.

In Fig. 2, we try to give a geometrical insight on the difference between FW and away steps in terms of search directions.

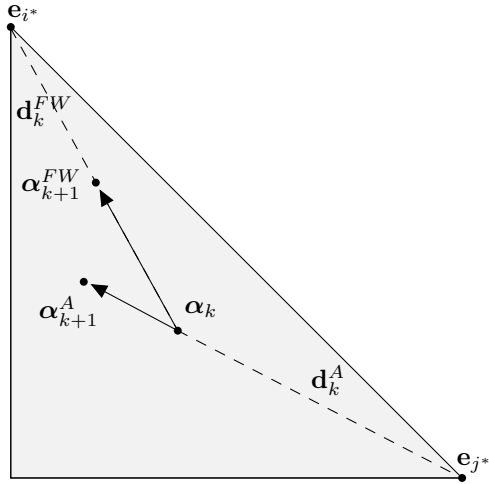
\begin{figure}[htb]
\begin{center}
\begin{tikzpicture}[line cap=round,line join=round,>=triangle 45,x=1.0cm,y=1.0cm, scale=1]\label{stepfigure}
\clip(-0.45,-0.3) rectangle (6.89,6.64);
\fill[color=black,fill=gray,fill opacity=0.1] (0,6) -- (0,0) -- (6,0) -- cycle;
\draw (6,0)-- (0,0);
\draw (0,0)-- (0,6);
\draw [color=black] (0,6)-- (0,0);
\draw [color=black] (0,0)-- (6,0);
\draw [color=black] (6,0)-- (0,6);
\draw [dash pattern=on 4pt off 4pt] (0,5.98)-- (2.22,1.96);
\draw [dash pattern=on 4pt off 4pt] (2.22,1.96)-- (6,0);
\draw [->,color=black] (2.22,1.96) -- (1.18,3.84);
\draw [->,color=black] (2.22,1.96) -- (1.06,2.58);
\fill [color=black] (0,6) circle (1.5pt);
\draw[color=black] (0.18,6.19) node {$\mathbf{e}_{i^{\ast}}$};
\fill [color=black] (6,0) circle (1.5pt);
\draw[color=black] (6.17,0.2) node {$\mathbf{e}_{j^{\ast}}$};
\fill [color=black] (2.22,1.96) circle (1.5pt);
\draw[color=black] (2.53,2.25) node {$\bm{\alpha}_k$};
\draw[color=black] (0.45,4.88) node {$\mathbf{d}_k^{FW}$};
\draw[color=black] (4.22,1.3) node {$\mathbf{d}_k^A$};
\fill [color=black] (1.13,3.93) circle (1.5pt);
\draw[color=black] (0.54,3.90) node {$\bm{\alpha}_{k+1}^{FW}$};
\fill [color=black] (0.97,2.61) circle (1.5pt);
\draw[color=black] (0.77,2.21) node {$\bm{\alpha}_{k+1}^A$};
\end{tikzpicture}
 \caption{A sketch of the search directions used by FW and MFW.}
\end{center}
\end{figure}

We previously hinted at the linear convergence properties of MFW. This result can now be stated more precisely \cite{yildirim08}.
\begin{proposition}
At each iteration of the MFW algorithm, we have:
\begin{equation}
\frac{\Phi(\bm{\alpha})- \Phi(\bm{\alpha}_{k+1})}{\Phi(\bm{\alpha})- \Phi(\bm{\alpha}_k)} \leq M \,,
\end{equation}
where $M \leq 1- \frac{1}{36m\beta d_S}$, $\beta$ is a constant and $d_S=\diam(S)^2$.
\end{proposition}

Substantially, as shown in the convergence analysis of
\cite{GuelatMarcotte}, there exists a point in the
optimization path of the MFW algorithm after which only away steps
are performed. That is, the algorithm only needs to remove useless
examples to correctly identify the optimal support vector set. From this stage on, the algorithm converges linearly to
the optimum value of the objective function. In contrast, the standard FW
algorithm does not possess the explicit ability to eliminate
spurious patterns from the index set, and tends to slow down when
getting near the solution.

\subsection{Beyond Normalized Kernels}

The methods studied in this paper were originally motivated by
recent advances in computational geometry that led to efficient
algorithms to address the MEB problem \cite{yildirim08}. However, a
major advantage of the proposed methods, over e.g. the CVM approach,
is that both the theory and the implementation of our algorithms can
be applied without changes to train SVMs using kernels which do not
satisfy condition (\ref{eq:hypothesis-kernel}), imposed to
obtain the equivalence between the MEB problem
(\ref{eq:ballProblemDual}) and the SVM optimization problem
(\ref{dualL2SVM}).

Both the FW and MFW methods were designed to maximize any
differentiable concave function $f(\bm{\alpha})$ in a bounded convex
polyhedron. The objective function in the SVM problem
(\ref{dualL2SVM}) is concave and the set of constraints coincides
with the unit simplex. The proposed methods can thus be applied
directly to solve (\ref{dualL2SVM}) without regard to 
(\ref{eq:ballProblemDual}). Theoretical results such as the global
convergence of algorithms still hold. In addition, since strict
complementarity usually holds for SVM problems, results in
\cite{GuelatMarcotte} imply that MFW still converges linearly to
the optimum. Note also that the constant $\Delta^{2}$, which makes
the difference between (\ref{eq:ballProblemDual}) and
(\ref{dualL2SVM}) for normalized kernels, can still be added to the
objective function of (\ref{dualL2SVM}) in the case of
non-normalized kernels, since it is simply ignored when optimizing for
$\bm{\alpha}$. An implementation designed to handle normalized
kernels can thus be directly used with any Mercer kernel.

It is apparent that the geometrical interpretation underlying
Algorithms \ref{alg:fvms} and \ref{alg:mfvms} needs to be
reformulated if the SVM problem is no longer equivalent to the
problem of computing a MEB. However, it is easy to show that the
search direction of the FW procedure at iteration $k$ is still
$\mathbf{e}_{i^{\ast}}$ where the index $i^{\ast}$ corresponds to
the largest component of $\nabla f(\bm{\alpha}_k)$. Similarly, the
away direction explored by MFW at iterate $k$ is still
$\mathbf{e}_{j^{\ast}}$ where the index $j^{\ast}$ corresponds to
the smallest component of $\nabla f(\bm{\alpha}_k)$. The set of
constraints in problem (\ref{dualL2SVM}) coincides with that of
(\ref{eq:ballProblemDual}). In addition, any approximate solution
$\bm{\alpha}_k$ produced by the proposed algorithms is feasible.
Thus, the sequence $f(\bm{\alpha}_k)$ is strictly increasing and converges
from below to the optimum
$f(\bm{\alpha}^{\ast})$. It is not immediately evident, however,
whether the stopping condition used within our algorithms guarantees
the method to find a solution in a neighborhood of the optimum
$\bm{\alpha}_k^{*}$. We now show that this is indeed the case. For simplicity of notation, it is convenient to write
explicitly the target QP of MEB-SVMs in matrix form:
\begin{equation}\label{dualL2SVM-matrix}
\begin{aligned}\maximize_{\bm{\alpha}} & \;\; g(\bm{\alpha}) := \Delta^{2} - \bm{\alpha}\tilde{\bm{K}}\bm{\alpha} = \Delta^{2} -  \|\bc\|^{2} \\
\mbox{subject to} & \;\; \mathbf{e}^T\bm{\alpha} = 1, \;\; \bm{\alpha}
\geq 0 \,,
\end{aligned}
\end{equation}
where $\tilde{\bm{K}}$ is the matrix of entries $\tilde{k}_{ij} =
\varphi(\bx_i)^{T}\varphi(\bx_j) = y_i y_j k(\bx_i,\bx_j) + y_i y_j
+ \frac{\delta_{ij}}{C}$, $\bc = \bm{Z}\bm{\alpha}$, and $\bm{Z}$ is
the matrix whose columns are the feature vectors $\bz_i =
\varphi(\bx_i)$. Note that $\bm{K} = \bm{Z}^{T}\bm{Z}$. When $k$ is a normalized kernel, we get
$\Phi(\bm{\alpha}) = g(\bm{\alpha})$. For
non-normalized kernels, instead, $\Delta^{2}$ can be viewed as an arbitrary
constant added to the SVM objective function in (\ref{dualL2SVM}),
$\Theta(\bm{\alpha}) = -\bm{\alpha}\tilde{\bm{K}}\bm{\alpha}$. That is, we can always think of $g(\bm{\alpha})$
as the objective function when solving (\ref{dualL2SVM}).

It is not hard to see that the stopping condition used in Algorithms
\ref{alg:fvms} and \ref{alg:mfvms} can be written as follows:
\begin{equation} \label{stopping-condition-1}
\begin{aligned}
                & \delta_{k+} \leq (1+\epsilon)^2 - 1\\
\Longleftrightarrow \,\,& \|\bz_{i^{\ast}} - \bc_k\|^{2} \leq (1+\epsilon)^2 r_k^{2}\\
\Longleftrightarrow \,\,& \Delta^{2} - 2\bz_{i^{\ast}}^{T}\bc_k +
\|\bc_k\|^{2} \leq (1+\epsilon)^2 r_k^{2} \, .
\end{aligned}
\end{equation}
Since by construction $r_k^{2} = \Delta^{2} - \|\bc_k\|^{2}$, we get
\begin{equation}\label{stopping-condition-2}
\begin{aligned}
& \Delta^{2} - 2\bz_{i^{\ast}}^{T}\bc_k + \|\bc_k\|^{2} \leq
(1+2\epsilon+\epsilon^2) \left(\Delta^{2} - \|\bc_k\|^{2}\right)\\
\Longleftrightarrow \,\,& \Delta^{2} - 2\bz_{i^{\ast}}^{T}\bc_k +
\|\bc_k\|^{2} \leq (1+\varepsilon)\left(\Delta^{2} - \|\bc_k\|^{2}\right)\\
\Longleftrightarrow \,\,& \Delta^{2} - 2\bz_{i^{\ast}}^{T}\bc_k +
\|\bc_k\|^{2} \leq \Delta^{2} - \|\bc_k\|^{2} + \varepsilon g(\bm{\alpha}_k)\\
\Longleftrightarrow \,\,& - 2\bz_{i^{\ast}}^{T}\bc_k + 2
\|\bc_k\|^{2} \leq \varepsilon g(\bm{\alpha}_k)\, ,
\end{aligned}
\end{equation}
with $\varepsilon = 2\epsilon+\epsilon^2 = \mathcal{O}(\epsilon)$.
%Note that the last equivalence follows from the fact that $g(\bm{\alpha}) =\Phi(\bm{\alpha})$. 
Now, since $\nabla g(\bm{\alpha}_k) = -2\bm{K}\bm{\alpha}_k = -2\bm{Z}^{T}\bc_k$, we
have that $\nabla g(\bm{\alpha}_k)_{i^{\ast}} =
-2\bz_{i^{\ast}}^{T}\bc_{k}$. In addition, $\bm{\alpha}_k^{T}\nabla
g(\bm{\alpha}_k) = -2\bm{\alpha}_k^{T}\bm{K}\bm{\alpha}_k =
-2\bm{\alpha}_k^{T}\bm{Z}^{T}\bm{Z}\bm{\alpha}_k =
-2\bc_{k}^{T}\bc_{k} = -2 \|\bc_{k}\|^{2}$. Thus, the stopping
condition for both algorithms is equivalent to
\begin{equation}
\nabla g(\bm{\alpha}_k)_{i^{\ast}} - \bm{\alpha}_k^{T}\nabla
g(\bm{\alpha}_k) \leq \varepsilon g(\bm{\alpha}_k) \, .
\label{stopping-condition-grads}
\end{equation}
On the other hand, since the objective function $g(\bm{\alpha})$ is
concave and differentiable,
\begin{equation}
g(\bm{\alpha}^{\ast}) \leq g(\bm{\alpha}_k) +
\left(\bm{\alpha}^{\ast} - \bm{\alpha}_k \right)^{T} \nabla
g(\bm{\alpha}_k) \, . \label{concave-property}
\end{equation}
In addition, $\mathbf{e}^T\bm{\alpha}^{\ast} = 1$ and thus
$\bm{\alpha}^{\ast T}\nabla g(\bm{\alpha}_k) \leq \max_{i \in \mathcal{I}} \nabla
g(\bm{\alpha}_k)_i = \nabla g(\bm{\alpha}_k)_{i^{\ast}}$. Therefore,
\begin{equation}\label{inequality-optimum}
\begin{aligned}
g(\bm{\alpha}^{\ast}) & \leq g(\bm{\alpha}_k) + \bm{\alpha}^{\ast T}\nabla
g(\bm{\alpha}_k)  - \bm{\alpha}_k^{T}
\nabla g(\bm{\alpha}_k)\\
& \leq g(\bm{\alpha}_k) + \nabla g(\bm{\alpha}_k)_{i^{\ast}} - \bm{\alpha}_k^{T}
\nabla g(\bm{\alpha}_k) \, .
\end{aligned}
\end{equation}
In virtue of (\ref{stopping-condition-grads}) and (\ref{inequality-optimum}), we obtain that
%the Algorithms (\ref{alg:fvms}) and (\ref{alg:mfvms}) stop if and only if
\begin{equation} \label{stopping-in-function-of-the-optimum}
\begin{aligned}
g(\bm{\alpha}^{\ast}) & \leq g(\bm{\alpha}_k) + \varepsilon
g(\bm{\alpha}_k) = (1+\varepsilon)g(\bm{\alpha}_k) \, .
\end{aligned}
\end{equation}
Finally, from the feasibility of $\bm{\alpha}_k$, we have
$g(\bm{\alpha}_k) \leq g(\bm{\alpha}^{\ast})$. Therefore,
\begin{equation}
\begin{aligned}
(1-\varepsilon) g(\bm{\alpha}^{\ast}) & \leq g(\bm{\alpha}_k) \leq
g(\bm{\alpha}^{\ast}) \, , \label{eq:correct-stop}
\end{aligned}
\end{equation}
that is, Algorithms \ref{alg:fvms} and \ref{alg:mfvms} stop with
an objective function value $g(\bm{\alpha}_k)$ in a left neighborhood of
radius $g(\bm{\alpha}^{\ast})\varepsilon = g(\bm{\alpha}^{\ast})(2\epsilon+\epsilon^2) = \mathcal{O}(\epsilon)$
around the optimum, even if the target problem (\ref{dualL2SVM}) is
not equivalent to a MEB problem.

\section{Experiments}

We test all the classifications methods discussed above on several classification problems. Our aim is to show that, as long as a minor loss in accuracy is acceptable, Frank-Wolfe based methods are able to build $L_2$-SVM classifiers in a considerably smaller time compared to CVMs, which
in turn have been proven in \cite{coreSVMs05tsang} to be faster than most traditional SVM software. This is especially evident on large-scale problems, where the capability to construct a classifier in a significantly reduced amount of time may be most useful. 

\subsection{Organization of this Section}
\label{subsec:org}
After discussing several implementation issues we compare the performance of the studied algorithms on several classical datasets. Our experiments include scalability tests on two different collections of problems of increasing size, which assess the capability of Frank-Wolfe based methods to efficiently solve problems of increasingly large size. These results can be found in Subsecs. \ref{subsec:web} and \ref{subsec:adult}. In Subsection \ref{subsec:single} we present additional experiments on the set of problems studied in  \cite{CIARP}. The statistical significance of the results presented so far is analyzed in section \ref{subsec:significance}.  A separate test is then performed in Subsection \ref{subsec:C} to study the influence of the penalty parameter $C$ on each training algorithm. Finally, in Subsection \ref{subsec:otherKernels} we present some experiments showing the capability of FW and MFW methods to handle a wider family of kernel function with respect to CVMs. We highlight that the purpose of that paragraph is not to improve the accuracy or the training time of the algorithms. A detailed commentary on the obtained results, which summarizes and expands on our conclusions, closes the section in Subsection \ref{subsec:discussion}.

\subsection{Datasets and Implementation Issues}\label{subsec:implementation}
As we detail below, all the datasets used in this section have been widely used in the literature. They were selected to cover a large variety with respect to the number of instances, number of dimensions and number of classes. In most of the cases, the training and testing sets are standard (precomputed for benchmarking) and can be obtained from public repositories like \cite{SVMLIB}, \cite{UCI2010}, or others we indicate in the dataset descriptions. The exceptions to this rule are the datasets \textbf{Pendigits} and \textbf{KDD99-10pc}. In these cases, the testing set was obtained by random sampling from the original collection a $20\%$ of the items. All the examples not selected as test instances were employed for training.  

For each problem, we specify, in Tab. \ref{TabDatasets}, the number $m$ of training points, the input space dimension $n$, and the number of classes $K$. We indicate by $t$ the number of examples in the {\em test set}, which is used to evaluate the accuracy of the classifiers but never employed for training or parameter tuning. In the case of multi-category classification problems, we adopted the one-versus-one approach (OVO), which is the method used in \cite{coreSVMs05tsang} to extend CVMs beyond binary classification and that usually obtains the best performance both in
terms of accuracy and training time according to \cite{Hsu02ComparisonMultiClassSVMs}. Hence, for these cases we also report the size $m_{{\max}}$ of the largest binary subproblem and the size $m_{{\min}}$ of the smallest binary subproblem in the OVO decomposition. 

%\rowcolors{1}{gray!40}{white}
\begin{table}[h]
\centering
\begin{small}
\begin{tabular}{|@{\,\,\,\,} p{2.4cm} | @{\,\,\,\,}p{1.6cm}@{\,\,}p{1.6cm}@{\,\,}p{1cm}@{\,\,}p{1.6cm}@{\,\,}p{1.6cm}@{\,\,}p{1cm}@{\,\,\,}|}
\hline
 % &   &  &  &  &   & \\
{Dataset} & ${m}$ & $t$ & ${K}$ & $m_{{\max}}$ & $m_{{\min}}$ & ${n}$\\
  % &   &  &  &  &   & \\
\hline
%{\textbf{Glass}}  & 119     & & 6   &   & 9\\
%{\textbf{Wine}}     & 142   & & 3   &   & 13\\
%{\textbf{Iris}}     & 171   & & 3   &   & 4\\
%   &   &  &  &  &   & \\
{\textbf{USPS}}   & 7291  & 2007 & 10  & 2199 & 1098  & 256\\
{\textbf{Pendigits}}     & 7494 & 3498 & 10   & 1560 & 1438  & 16\\
{\textbf{Letter}}    & 15000    & 5000 & 26   & 1213 & 1081  & 16\\
{\textbf{Protein}}  & 17766  & 6621 & 3  & 13701 & 9568 & 357\\
{\textbf{Shuttle}}   & 43500    & 14500  & 7    & 40856 & 17 & 9\\
{\textbf{IJCNN}}     & 49990     & 91701 & 2   & 49990 & 49990  & 22\\
{\textbf{MNIST}}     & 60000    & 10000 & 10   & 13007 & 11263  & 780\\
{\textbf{USPS-Ext}}  & 266079    & 75383 & 2   & 266079 & 266079 & 676\\
{\textbf{KDD99-10pc}}    & 395216   & 98805 & 5  & 390901 & 976 & 127\\
{\textbf{KDD99-Full}}    & 4898431   & 311029 & 2   & 4898431 & 4898431 & 127\\
{\textbf{Reuters}}   & 7770 & 3299 & 2    & 7770 & 7770 & 8315\\
{\textbf{Adult a1a}}     & 1605 & 30956 & 2    &  1605 & 1605 & 123\\
{\textbf{Adult a2a}}     & 2265 & 30296 & 2    &  2265 & 2265 & 123\\
{\textbf{Adult a3a}}     & 3185 & 29376 &  2    &  3185 & 3185 & 123\\
{\textbf{Adult a4a}}     & 4781 & 27780   & 2    &  4781 & 4781 & 123\\
{\textbf{Adult a5a}}     & 6414 & 26147 & 2    &  6414 & 6414 & 123\\
{\textbf{Adult a6a}}     & 11220    & 21341   & 2    & 11220 & 11220  & 123\\
{\textbf{Adult a7a}}     & 16100    & 16461  & 2    & 16100 & 16100  & 123\\
%{\textbf{Adult a8a}}     & 22696    & 9865 & 2    & 22696 & 22696  & 123\\
{\textbf{Web w1a}}   & 2477 & 47272 & 2    & 2477 & 2477 & 300\\
{\textbf{Web w2a}}   & 3470 & 46279 & 2    & 3470 & 3470 & 300\\
{\textbf{Web w3a}}   & 4912 & 44837 & 2    & 4912 & 4912 & 300\\
{\textbf{Web w4a}}   & 7366 & 42383 & 2    & 7366 & 7366 & 300\\
{\textbf{Web w5a}}   & 9888 & 39861 & 2    & 9888 & 9888 & 300\\
{\textbf{Web w6a}}   & 17188    & 32561 & 2    & 17188 & 17188 & 300\\
{\textbf{Web w7a}}   & 24692    & 25057 & 2    & 24692 & 24692 & 300\\
{\textbf{Web w8a}}   & 49749    & 14951  & 2    & 49749 & 49749 & 300\\
%   &   &  &  &  &   & \\
\hline
\end{tabular}
\end{small} \caption{\label{TabDatasets} Features of the selected
datasets.}
\end{table}
Here follow some brief descriptions of the pattern recognition problems underlying each dataset, taken from their respective sources.
\begin{itemize}
\item \textbf{USPS}, \textbf{USPS-Ext} - The USPS dataset is a classic handwritten digits recognition problem, where the patterns are $16\times 16$ images from United States Postal Service envelopes. The extended version USPS-Ext first appeared in \cite{coreSVMs05tsang} to show the large-scale capabilities of CVMs. The original version can be downloaded from \cite{SVMLIB} and the extended one from \cite{LibCVM09}.

\item \textbf{Pendigits} - Another digit recognition dataset, created by experimentally collecting samples from a total of 44 writers with a tablet and a stylus. This dataset can be obtained from \cite{UCI2010}. 

\item \textbf{Letter} - An Optical Character Recognition (OCR) problem. The objective is to identify each of a large number of black-and-white rectangular pixel displays as one of the 26 capital letters in the English alphabet. The files can be obtained from \cite{SVMLIB}. 

\item \textbf{Protein} - A bioinformatics problem regarding protein structure prediction. This dataset can be download from \cite{SVMLIB}. 

\item \textbf{Shuttle} - This is a dataset in the Statlog collection, originated from NASA and concerning the position of radiators within the Space Shuttle \cite{MST}. The dataset can be obtained from \cite{UCI2010} or \cite{LibCVM09}. 

\item \textbf{IJCNN} - A dataset from the 2001 neural network competition of the International Joint Conference on Neural Networks. We obtained this dataset from \cite{coreSVMs05tsang}. 

\item \textbf{MNIST} - Another classic handwritten digit recognition problem, this time coming from National Institute of Standards (NIST) data. The dataset can be obtained from \cite{LibCVM09}.

\item \textbf{KDD99-10pc}, \textbf{KDD99-Full} - This is a dataset used in the 1999 Knowledge Discovery and Data Mining Cup. The data are connection records for a network,  obtained by simulating a wide variety of normal accesses and intrusions on a military network. The problem is to detect different types of accesses on the network with the aim of identifying fraudulent ones. The 10pc version is a randomly selected $10\%$ of the whole data.

\item \textbf{Reuters} - A text categorization problem built from a collection of documents that appeared on Reuters newswire in 1987. The documents were assembled and indexed with categories. The binary version used in this paper (relevant versus non-relevant documents) was obtained from \cite{LibCVM09}.

\item \textbf{Adult a1a-a8a} - A series of problems derived from a dataset extracted from the 1994 US Census database. %, also known as ``Census Income'' dataset.
The original aim was to predict whether an individual's income exceeded $50000$US$\$$/year, based on personal data. All the instances of this collection can be downloaded from \cite{LibCVM09}. 

\item \textbf{Web w1a-w8a} - A series of problems extracted from a web classification task dataset, first appeared in Platt's paper on Sequential Minimal Optimization for training SVMs \cite{platt99smo-seminal}. All the instances of this collection can be downloaded from \cite{LibCVM09}. 
\end{itemize}

\subsubsection{SVM Parameters}
For the experiments presented in Subsection \ref{subsec:web} to Subsection \ref{subsec:C}, SVMs were trained using a RBF kernel $k(\mathbf{x}_{1},\mathbf{x}_{2}) = \exp(-\|\mathbf{x}_{1}-\mathbf{x}_{2}\|^{2}/2 \sigma^{2})$. The reason for this choice is that this kernel is the best-known in the family of kernels admitted by CVMs and it is frequently used in practice \cite{coreSVMs05tsang}. In particular, this is the choice for the large set of experiments presented in \cite{coreSVMs05tsang} to demonstrate the advantage of this framework on other SVM software. However, in Subsection \ref{subsec:otherKernels} we present some results showing the capability of FW and MFW methods to handle a polynomial kernel, which does not satisfy the conditions required by CVMs.

For the relatively small datasets \textbf{Pendigits} and \textbf{USPS}, parameter $\sigma^{2}$ was determined together with parameter $C$ of SVMs using $10$-fold cross-validation on the logarithmic grid $[2^{-15},2^{5}] \times [2^{-5},2^{15}]$, where the first collection of values correspond to parameter $\sigma^{2}$ and the second to parameter $C$. For the large-scale datasets, $\sigma^{2}$ was determined using the default method employed by CVMs in \cite{coreSVMs05tsang}, that is, it was set to the average squared distance among training patterns. Parameter $C$ was determined on the logarithmic grid $[2^0,2^{12}]$ using a validation set consisting in a randomly computed 30\% fraction of the training set. 

We stress that the aim of this paper is not to determine an optimal value of the parameters by fine-tuning each algorithm on the test problems to seek for the best possible accuracy. As our intent is to compare the performance of the presented methods and analyze their behavior in a manner consistent with our theoretical analysis, it is necessary to perform the experiments under the same conditions on a given dataset. That is to say, the optimization problem to be solved should be the same for each algorithm. For this reason, we deliberately avoided using different training parameters when comparing different methods.  Specifically, parameters $\sigma^{2}$ and $C$ were tuned using the CVM method and the obtained values were used for all the algorithms discussed in this paper (CVM, FW and MFW). 

Furthermore, since the value of parameter $C$ can have a significant influence on the running times, we devote a specific subsection to
evaluate the effect of this parameter on the different training algorithms.

 \subsubsection{MEB Initialization and Parameters}
As regards the initialization of the CVM, FW and MFW methods, that is, the computation of $\mathcal{I}_{0}$ and $\bm{\alpha}_{0}$ in
Algorithms \ref{alg:cvms}, \ref{alg:fvms} and \ref{alg:mfvms}, we adopted the random MEB method described in the previous sections,
using $p=20$ points. As suggested in \cite{coreSVMs05tsang}, we used $\epsilon=10^{-6}$ with all the algorithms.

\subsubsection{Random Sampling Techniques}

Computing $i^{\ast}$, i.e. evaluating (\ref{distance}) for all of
the $m$ training points, requires a number of kernel evaluations of
order $\mathcal{O}(q_k^2+mq_k) = \mathcal{O}(mq_k)$, where $q_k$ is the cardinality of
$\mathcal{I}_k$. If $m$ is very large, this complexity can quickly
become unacceptable, ruling out the possibility to solve large scale
classification problems in a reasonable time. A sampling technique,
called \emph{probabilistic speedup}, was proposed in
\cite{SS2} to overcome this obstacle. In practice, the distance
(\ref{distance}) is computed just on a random subset $\varphi(S^{\prime})
\subset \varphi(S)$, where $S^{\prime}$ is identified by an index set $\mathcal{I}^{\prime}$ of small constant cardinality $r$. The overall complexity
is thereby reduced to order $\mathcal{O}(q_k^2 + q_k) = \mathcal{O}(q_k^2)$, a major
improvement on the previous estimate, since we generally have $q_k \ll
m$. The main result this technique relies on is the following
\cite{Scholkopf01Kernels}.
\begin{theorem}
Let $D:=\{d_1,\ldots,d_m\} \subset \mathbb{R}$ be a set of
cardinality $m$, and let $D^{\prime} \subset S$ be a random subset
of size $r$. Then the probability that $\max D^{\prime}$ is greater
or equal than $\tilde m$ elements of $D$ is at least $1-(\frac{\tilde m}{m})^r$.
\end{theorem}
For example, if $r = 59$ and $\tilde m=0.95m$,then with probability at
least $0.95$ the point in $\varphi(S^{\prime})$ farthest from the center lies among
the $5\%$ of the farthest points in the whole set $\varphi(S)$. This is the
choice originally made in \cite{coreSVMs05tsang} and used in \cite{CIARP} to test the CVM and FW algorithms.

\subsubsection{Caching}

We also adopted the LRR caching strategy designed in 
\cite{LibCVM09} for CVMs to avoid the computation of recently
used kernel values.

\setlength\floatsep{0pt} \setlength\intextsep{0pt}
\setlength{\abovecaptionskip}{0pt}
\setlength{\belowcaptionskip}{0pt} \setlength{\tabcolsep}{0pt}

\begin{figure}[ht]
\centering
\begin{tabular}{c}
\includegraphics[width=\textwidth, height=0.22\textheight]{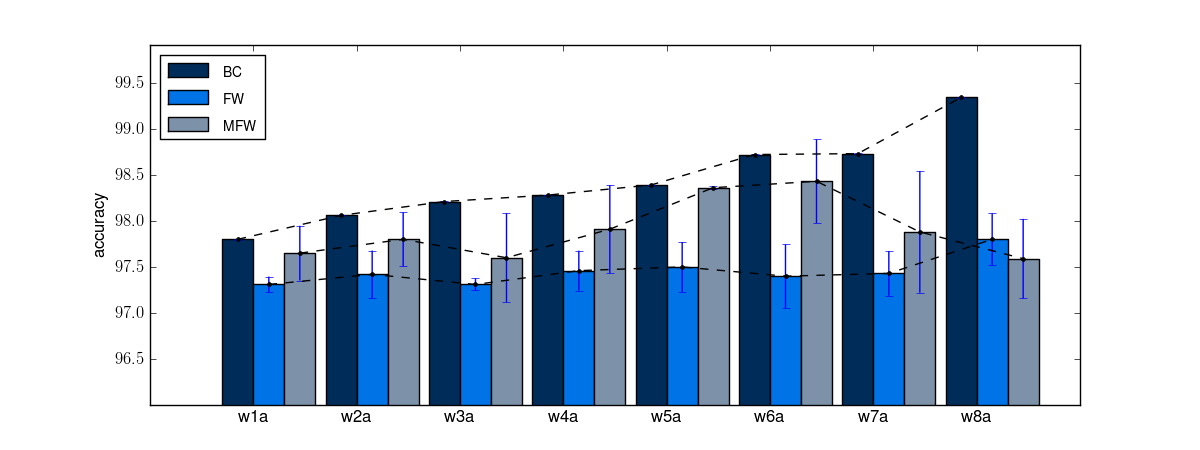}\\
\includegraphics[width=\textwidth, height=0.22\textheight]{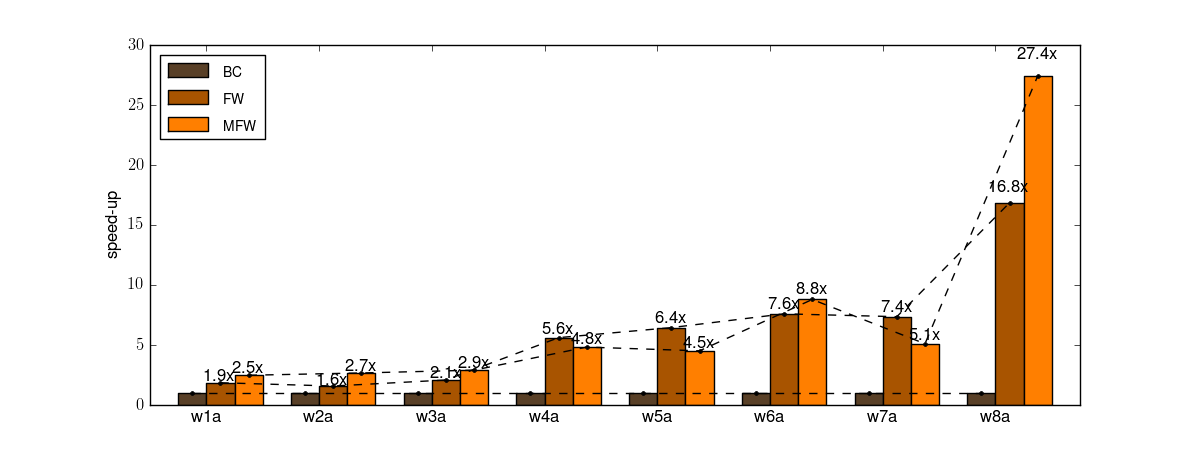}\\
\includegraphics[width=\textwidth, height=0.22\textheight]{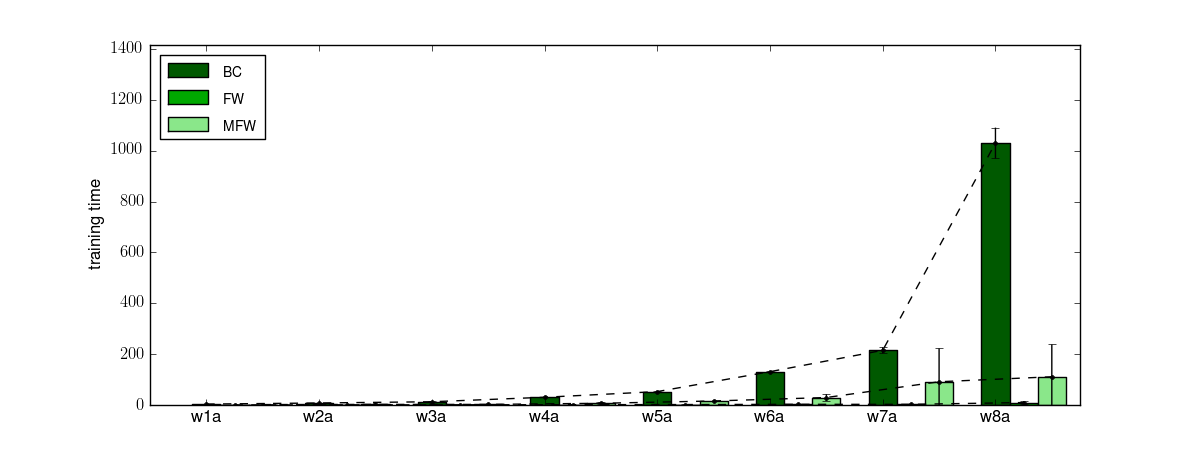}\\
\includegraphics[width=\textwidth, height=0.22\textheight]{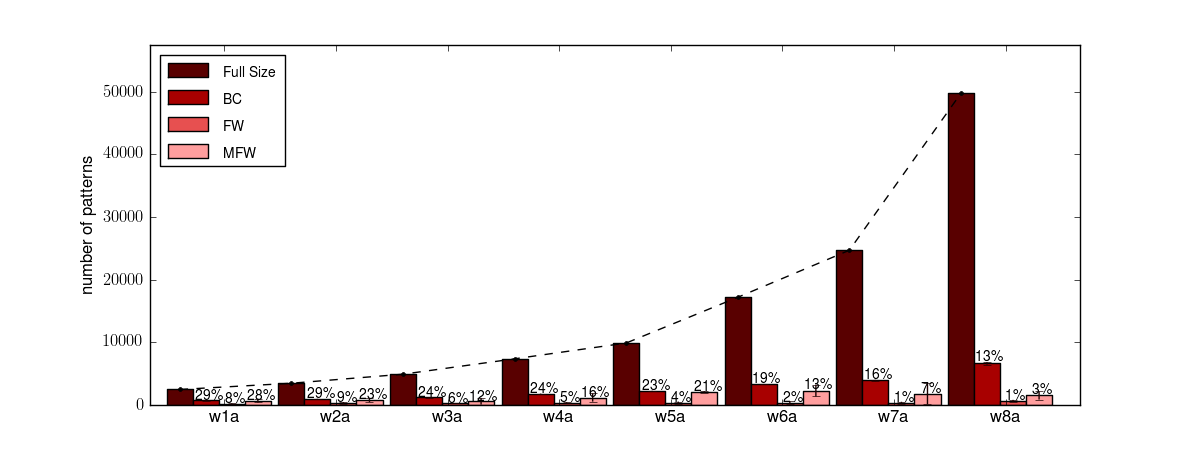}
\end{tabular}
\caption{Comparison of accuracies (first row), speed-ups (second row),
absolute running times (third row) and sizes of training sets and
support vector sets (fourth row) in the \textbf{Web} datasets.}
\label{webresults}
\end{figure}

\subsubsection{Computational Environment}
The experiments were conducted on a personal computer with a 2.66GHz Quad Core CPU and 4 GB of RAM, running 64bit GNU/Linux (Ubuntu 10.10). The algorithms were implemented based on the (C++) source code available at \cite{LibCVM09}.    

\subsection{Scalability Experiments on the Web Dataset Collection}
\label{subsec:web}

In Fig. \ref{webresults}, we report some results concerning
accuracies, training times, speed-ups and support vector set sizes obtained
in the \textbf{Web} datasets. The series is monotonically increasing
in the number of training patterns, which grows approximately as
$m_{i}=1.4^{i}m_{0}$, $i=1,\ldots,8$, where $m_0$ is the number of training patterns
in the first dataset \cite{platt99smo-seminal}.

The speed-up of the FW method with respect to CVMs is measured as
$s=t_{0}/t_{1}$ where $t_{0}$ is the training time of the CVM
algorithm and $t_1$ is the training time of the FW method, both
measured in seconds. Similarly, the speed-up of the MFW method with
respect to CVMs is measured as $s=t_{0}/t_{2}$ where $t_2$ is the
training time of the MFW method.

As depicted in Fig. \ref{webresults} the proposed methods are
slightly less accurate than CVMs. The training time, in contrast,
scales considerably better for our methods as the number of training
patterns increases. The speed-ups are actually always greater than
$1$, which shows that the FW and MFW methods indeed build
classifiers faster than CVMs. More importantly, the speed-up is
monotonically increasing, ranging from $12$ times faster up to $107$
times faster in the case of the FW algorithm and from $2$ times
faster up to almost $10$ times faster in the case of the MFW method.
This suggests that the improvements of the proposed method over CVMs
becomes more and more significant as the size of the training set
grows.

\subsection{Scalability Experiments on the Adult Dataset Collection}
\label{subsec:adult}

\begin{figure}[h!!!]
\centering
\begin{tabular}{c}
\includegraphics[width=\textwidth, height=0.22\textheight]{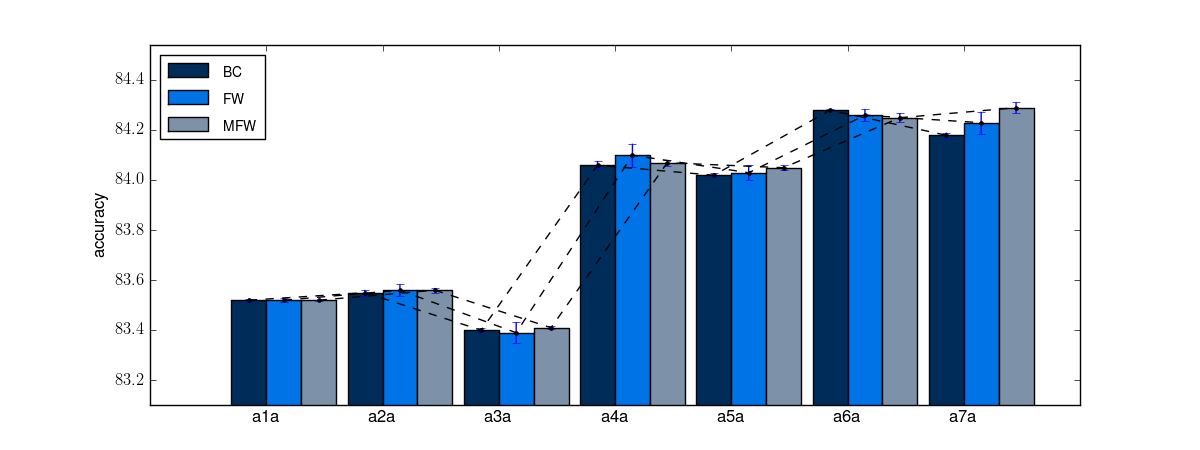}\\
\includegraphics[width=\textwidth, height=0.22\textheight]{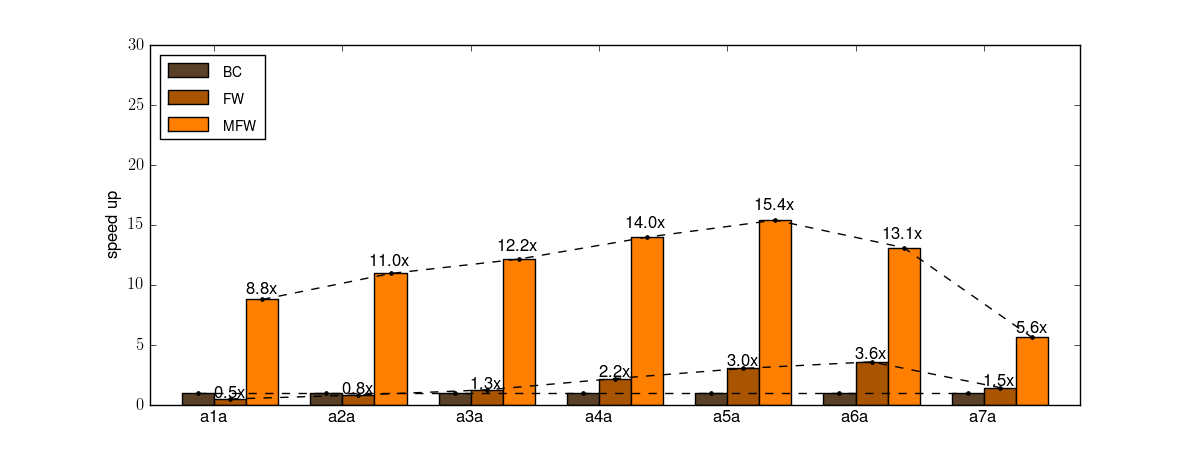}\\
\includegraphics[width=\textwidth, height=0.22\textheight]{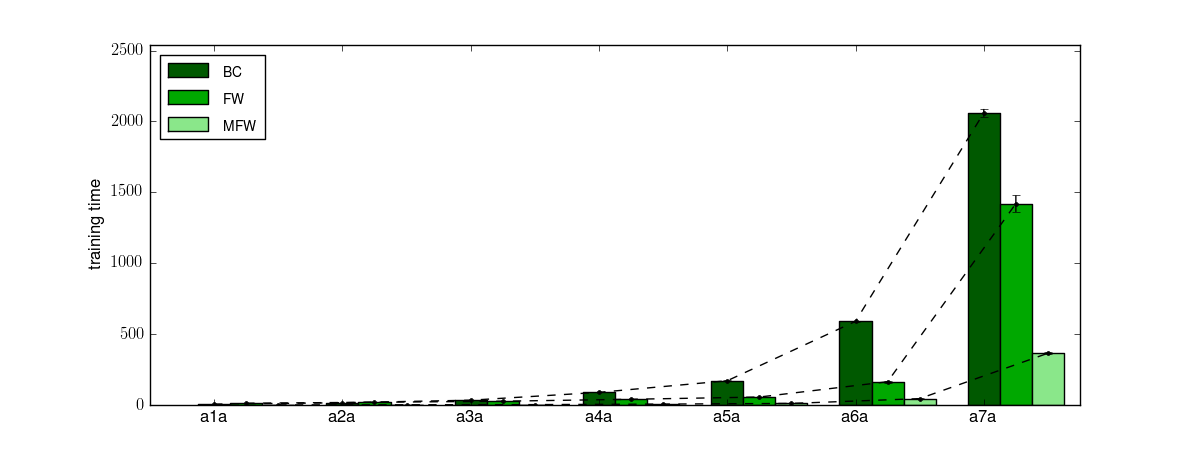}\\
\includegraphics[width=\textwidth, height=0.22\textheight]{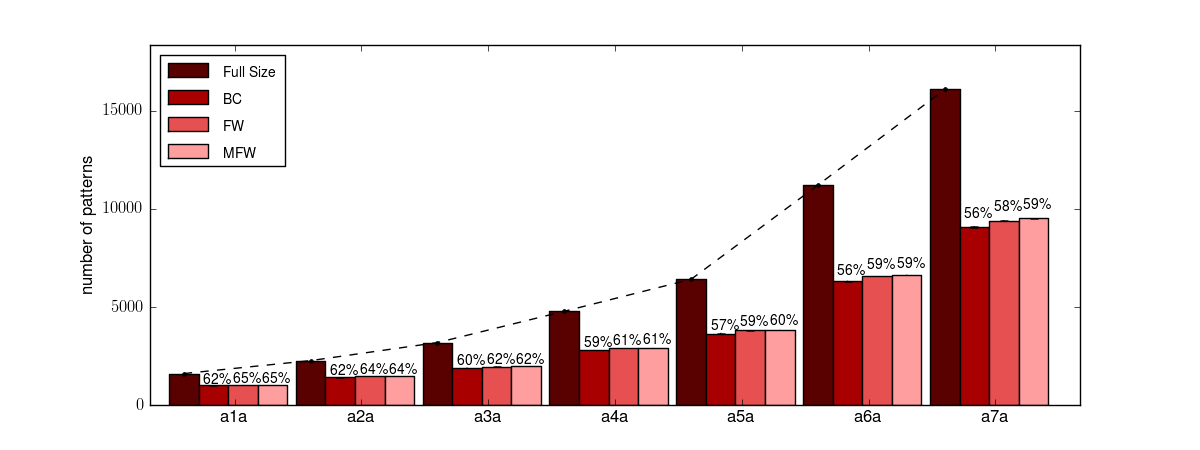}
\end{tabular}
\caption{Comparison of accuracies (first row), speed-ups (second row),
absolute running times (third row) and sizes of training sets and
support vector sets (fourth row) in the \textbf{Adult}
datasets.}\label{adultresults}
\end{figure}

Fig. \ref{adultresults} depicts accuracies and speed-ups
obtained in the \textbf{Adult} datasets. Like the \textbf{Web}
datasets, this collection was created with the purpose of analyzing
the scalability of SVM methods and the number of training patterns
grows approximately with the same rate \cite{platt99smo-seminal}. The
speed-up of the FW and MFW methods is computed as in the previous
section.

Results obtained in this experiment confirm that the proposed
methods tend to be faster than CVMs as the dataset grows. CVMs are
actually faster than FW just in two cases, corresponding to the
smallest versions of the sequence. MFW however runs always faster
than CVMs, reaching a speed-up of $15\times$ in the fifth version of
the series. The speed-ups obtained by the FW method are in this
experiment more moderate than in the \textbf{Web} collection.
However, most of the time FW exhibits also better test accuracies
than CVMs. Finally, the MFW algorithm is not only faster but also as
accurate as CVMs on this classification problem.

\subsection{Experiments on Single Datasets}\label{singledatasetsec}
\label{subsec:single}

\begin{figure}[h!!!]
\centering
\begin{tabular}{c}
\includegraphics[width=\textwidth, height=0.25\textheight]{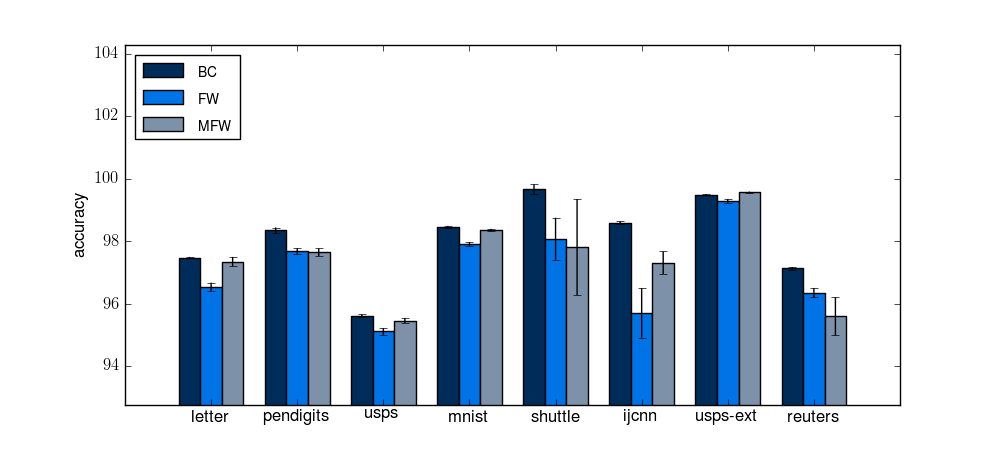}\\
\includegraphics[width=\textwidth, height=0.25\textheight]{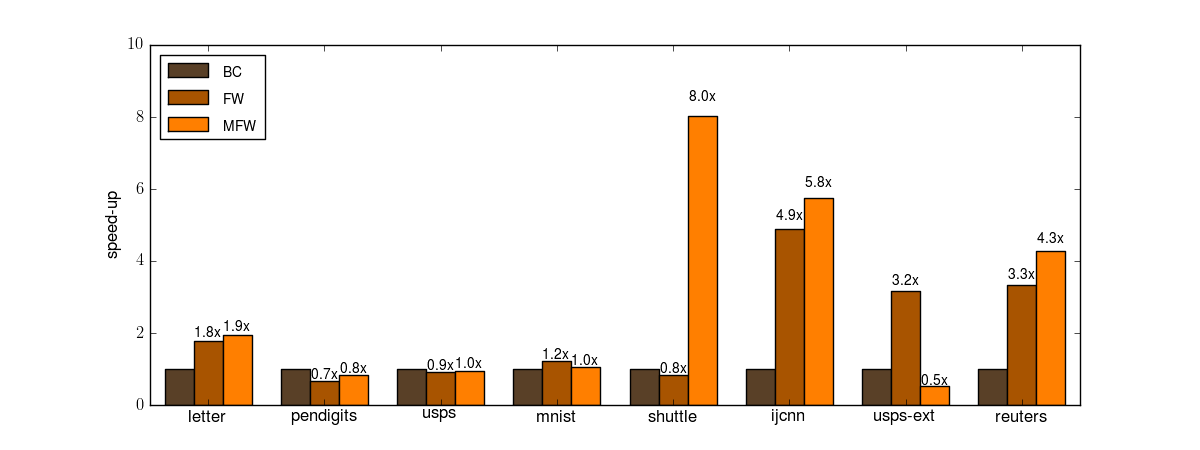}
\end{tabular}
\caption{Comparison of accuracies (first row) and speed-ups (second row) obtained in some of the single datasets of Tab. \ref{TabDatasets}}
\label{fig:comparison_single_1_of_2}
\end{figure}

\begin{figure}[h!!!]
\centering
\begin{tabular}{cc}
\includegraphics[width=0.5\textwidth, height=0.24\textheight]{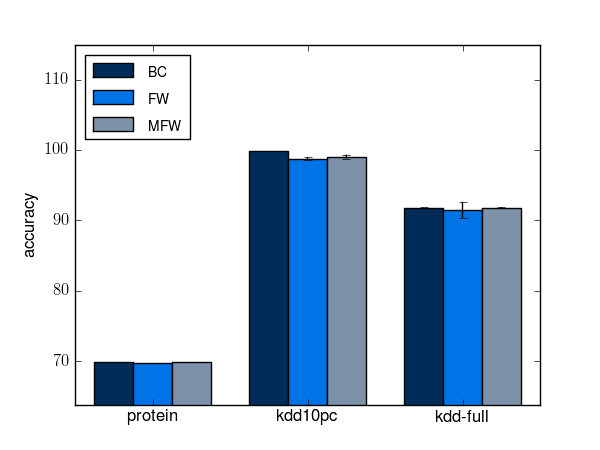} &
\includegraphics[width=0.5\textwidth, height=0.24\textheight]{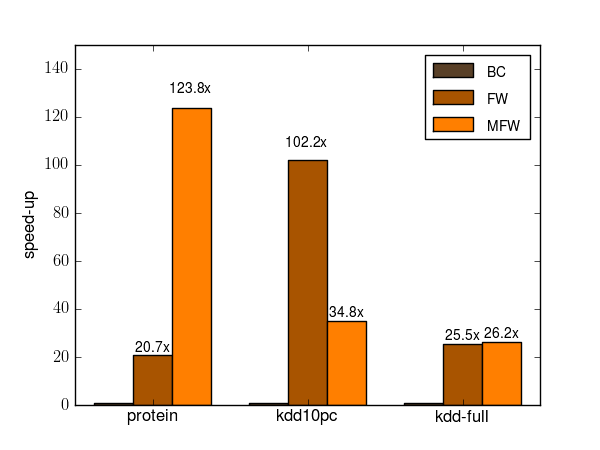}
\end{tabular}
\caption{Comparison of accuracies (first column) and speed-ups (second column) obtained in some of the single datasets of Tab. \ref{TabDatasets}}\label{fig:comparison_single_2_of_2}
\end{figure}

Results of Figs. \ref{fig:comparison_single_1_of_2} and
\ref{fig:comparison_single_2_of_2} correspond to accuracies and
speed-ups obtained in the single datasets described in Tab.
\ref{TabDatasets}, that is, all of them but the \textbf{Web} and
\textbf{Adult} series. Most of these problems have been already used
to show the improvements of CVMs over other algorithms to train
SVMs . 
%Tab. \ref{tab:summary1} in Appendix 1 additionally
%shows the absolute training times and testing accuracies in a
%tabular format in order to facilitate numerical comparisons.

Results show that the proposed methods are faster than CVMs most of
the time, sometimes at the price of a slightly lower accuracy. The
speed-up achieved by FW and MFW becomes more significant as the size
of the training set grows. FW in particular reaches peaks of
$102\times$ in the second largest dataset (\textbf{KDD99-10pc}) and
$25\times$ in the largest of the problems studied in this experiment
(\textbf{KDD99-Full}). Finally, the results show that in the cases for which
CVMs are faster, the advantage of this algorithm on FW and MFW tend
to be very small, reaching its better speed-up against MFW in the
\textbf{USPS-Ext} dataset, for which however FW exhibits a speed up of around
$3 \times$.

\vspace{1.0cm}

\newcolumntype{C}[1]{>{\centering\let\newline\\\arraybackslash\hspace{0pt}}m{#1}}
\newcolumntype{L}[1]{>{\raggedright\let\newline\\\arraybackslash\hspace{0pt}}m{#1}}
\newcolumntype{R}[1]{>{\raggedleft\let\newline\\\arraybackslash\hspace{0pt}}m{#1}}

\begin{table}[ht]\centering
\begin{small}
\begin{tabular}{|@{\,\,\,\,} p{1.5cm} @{\,\,\,\,} |@{\,\,\,\,} L{1.1cm}@{\,\,\,\,}|@{\,\,\,\,} L{1.4cm}@{\,\,\,\,}|@{\,\,\,\,} L{1.1cm}@{\,\,\,\,}  |@{\,\,\,\,} L{1.4cm}@{\,\,\,\,}|@{\,\,\,\,} L{1.1cm}@{\,\,\,\,}|@{\,\,\,\,} L{1.4cm}@{\,\,\,\,}|}
\hline
%& \multicolumn{6}{l|}{}\\   
Dataset &  \multicolumn{2}{c}{\textbf{CVM}} & \multicolumn{2}{c}{\textbf{FW}}& \multicolumn{2}{c|}{\textbf{MFW}}\\  
\cline{1-7}
& Acc.  & STD. & Acc. & STD. & Acc. & STD. \\ 
\hline
a1a&83.52&6.33E-003&83.52&7.53E-003&83.52&1.58E-003\\
a2a&83.55&1.15E-002&\underline{\textbf{83.56}}&2.39E-002&\underline{\textbf{83.56}}&9.19E-003\\
a3a&83.40&8.45E-003&83.39&4.17E-002&\underline{\textbf{83.41}}&5.00E-003\\
a4a&84.06&1.59E-002&\underline{\textbf{84.10}}&4.64E-002&84.07&1.15E-002\\
a5a&84.02&8.92E-003&84.03&2.93E-002&\underline{\textbf{84.05}}&9.86E-003\\
a6a&\underline{\textbf{84.28}}&3.51E-003&84.26&2.49E-002&84.25&1.86E-002\\
a7a&84.18&1.06E-002&84.23&4.41E-002&\underline{\textbf{84.29}}&2.26E-002\\
w1a&\underline{\textbf{97.80}}&2.81E-003&97.31&8.54E-002&97.65&2.97E-001\\
w2a&\underline{\textbf{98.06}}&1.37E-003&97.42&2.58E-001&97.80&2.93E-001\\
w3a&\underline{\textbf{98.21}}&1.67E-003&97.31&6.48E-002&97.60&4.83E-001\\
w4a&\underline{\textbf{98.28}}&9.44E-004&97.46&2.17E-001&97.91&4.79E-001\\
w5a&\underline{\textbf{98.39}}&2.01E-003&97.50&2.71E-001&98.36&1.76E-002\\
w6a&\underline{\textbf{98.72}}&5.63E-003&97.40&3.45E-001&98.43&4.54E-001\\
w7a&\underline{\textbf{98.73}}&5.29E-003&97.43&2.43E-001&97.88&6.60E-001\\
w8a&\underline{\textbf{99.34}}&5.35E-003&97.80&2.82E-001&97.59&4.32E-001\\
Letter&\underline{\textbf{97.48}}&2.19E-002&96.54&1.37E-001&97.35&1.50E-001\\
Pendigits&\underline{\textbf{98.35}}&9.46E-002&97.68&9.39E-002&97.65&1.22E-001\\
USPS&\underline{\textbf{95.63}}&3.73E-002&95.12&1.05E-001&95.47&8.34E-002\\
Reuters&\underline{\textbf{97.10}}&4.11E-002&96.40&1.53E-001&95.60&6.13E-001\\
MNIST&\underline{\textbf{98.46}}&3.14E-002&97.91&5.99E-002&98.36&4.05E-002\\
Protein&\underline{\textbf{69.79}}&0.00E+00&69.73&0.00E+00&69.78&0.00E+00\\
Shuttle&\underline{\textbf{99.67}}&1.51E-001&98.08&6.74E-001&97.82&1.54E+00\\
IJCNN&\underline{\textbf{98.59}}&4.89E-002&95.71&7.95E-001&97.31&3.63E-001\\
USPS-Ext&99.50&1.26E-002&99.30&5.47E-002&\underline{\textbf{99.57}}&2.76E-002\\
KDD10pc&\underline{\textbf{99.87}}&2.06E-002&98.82&2.13E-001&99.10&2.86E-001\\
KDD-full&91.77&7.17E-002&91.53&1.14E+00&\underline{\textbf{91.82}}&7.72E-002\\
%&&&&&&\\
\hline
\end{tabular}
\vspace{0.2cm}
\end{small}\caption{\label{table:accuracy_all} Test accuracy ($\%$) of the proposed algorithms and the baseline method CVM. Statistics correspond to the mean (Acc) and standard deviation (STD) obtained from $5$ repetitions of each experiment. For the Protein dataset, just $1$ repetition was carried out due to the significantly longer training times.}
\end{table}

\vspace{0.8cm}

\begin{table}[ht]\centering
\begin{small}
\begin{tabular}{|@{\,\,\,\,} p{1.5cm} @{\,\,\,\,} |@{\,\,\,\,} R{1.1cm}@{\,\,\,\,}|@{\,\,\,\,} R{1.4cm}@{\,\,\,\,}|@{\,\,\,\,} R{1.1cm}@{\,\,\,\,}  |@{\,\,\,\,} R{1.4cm}@{\,\,\,\,}|@{\,\,\,\,} R{1.1cm}@{\,\,\,\,}|@{\,\,\,\,} R{1.4cm}@{\,\,\,\,}|}
\hline
%& \multicolumn{6}{l|}{}\\   
Dataset &  \multicolumn{2}{c}{\textbf{CVM}} & \multicolumn{2}{c}{\textbf{FW}}& \multicolumn{2}{c|}{\textbf{MFW}}\\  
\cline{1-7}
& Time  & STD. & Time & STD. & Time & STD. \\ 
\hline
a1a&	6.26&	8.82E-02&	12.5&	6.99E-01&	\underline{\textbf{0.712}}&	4.00E-03\\
a2a&	16&	1.82E-01&	19.3&	1.51E+00&	\underline{\textbf{1.46}}&	1.41E-02\\
a3a&	33.8&	1.74E-01&	26.8&	1.44E+00&	\underline{\textbf{2.78}}&	6.53E-02\\
a4a&	89.1&	2.67E-01&	40.4&	1.19E+00&	\underline{\textbf{6.37}}&	1.08E-01\\
a5a&	171&	2.10E+00&	56.1&	1.94E+00&	\underline{\textbf{11.1}}&	1.75E-01\\
a6a&	590&	2.62E+00&	164&	6.24E+00&	\underline{\textbf{45}}&	1.15E+00\\
a7a&	2060&	2.79E+01&	1420&	5.86E+01&	\underline{\textbf{365}}&	7.30E+00\\
w1a&	3.59&	3.62E-01&	\underline{\textbf{0.286}}&	8.26E-02&	1.67&	7.62E-01\\
w2a&	7.9&	1.62E-01&	\underline{\textbf{0.658}}&	3.32E-01&	2.31&	1.20E+00\\
w3a&	12.8&	1.11E+00&	\underline{\textbf{0.76}}&	5.90E-02&	2.27&	2.25E+00\\
w4a&	30.4&	7.50E-01&	\underline{\textbf{1.26}}&	5.01E-01&	6.76&	4.13E+00\\
w5a&	52.7&	6.39E-01&	\underline{\textbf{1.78}}&	1.07E+00&	15.8&	1.44E+00\\
w6a&	131&	1.55E+00&	\underline{\textbf{2.67}}&	2.36E+00&	29.4&	1.31E+01\\
w7a&	215&	1.17E+01&	\underline{\textbf{2.58}}&	1.89E+00&	91.4&	1.31E+02\\
w8a&	1030&	6.05E+01&	\underline{\textbf{9.64}}&	5.22E+00&	111&	1.27E+02\\
Letter&	23.7&	2.80E-01&	13.3&	2.05E-01&	\underline{\textbf{12.3}}&	1.42E-01\\
Pendigits&	\underline{\textbf{0.554}}&	3.26E-02&	0.82&	2.97E-02&	0.658&	2.23E-02\\
USPS&	6.89&	7.46E-02&	7.58&	1.42E-01&	\underline{\textbf{7.22}}&	9.00E-02\\
Reuters&	7.24&	3.62E-01&	2.17&	3.87E-01&	\underline{\textbf{1.69}}&	5.87E-01\\
MNIST&	364&	1.31E+01&	\underline{\textbf{301}}&	8.56E+00&	349&	2.59E+00\\
Protein&	247000&	0.00E+00&	11900&	0.00E+00&	\underline{\textbf{2000}}&	0.00E+00\\
Shuttle&	1.41&	3.41E-01&	1.69&	4.56E-01&	\underline{\textbf{0.176}}&	2.73E-02\\
IJCNN&	198&	1.36E+01&	40.5&	2.27E+01&	\underline{\textbf{34.4}}&	1.26E+01\\
USPS-Ext&	84.4&	2.02E+01&	\underline{\textbf{26.7}}&	3.74E+00&	161&	1.49E+01\\
KDD10pc&	42.3&	3.58E+00&	\underline{\textbf{0.414}}&	1.50E-02&	1.22&	1.24E+00\\
KDD-full&	19.5&	8.12E+00&	0.764&	2.42E-02&	\underline{\textbf{0.744}}&	8.00E-03\\
%&&&&&&\\
\hline
\end{tabular}
\end{small}
\vspace{0.2cm}
\caption{\label{table:time_all} Running times (seconds) of the proposed algorithms and the baseline method CVM. Statistics correspond to the mean (Time) and standard deviation (STD) obtained from $5$ repetitions of each experiment. For the Protein dataset, just $1$ repetition was carried out due to the significantly longer training times. }
\end{table}
\vspace{0.6cm}
For sake of readability we include in Tab. \ref{table:accuracy_all} and Tab. \ref{table:time_all} a summary of the test accuracy and running times used to build the Figs. \ref{webresults} to \ref{fig:comparison_single_2_of_2}.
\vspace{0.4cm}

\setlength\floatsep{0pt} \setlength\intextsep{0pt}
\setlength{\abovecaptionskip}{0pt}
\setlength{\belowcaptionskip}{1pt} \setlength{\tabcolsep}{0pt}

\cleardoublepage

\subsection{Statistical Tests}
\label{subsec:significance}

\setlength\floatsep{5pt} \setlength\intextsep{6pt}
\setlength{\abovecaptionskip}{1pt}
\setlength{\belowcaptionskip}{1pt} \setlength{\tabcolsep}{4pt}

This paragraph is devoted to verify the statistical significance of the results obtained above. To this end we adopt the guidelines suggested in \cite{Demsar06Statistical}, that is, we first conduct a multiple test to determine whether the hypothesis that all the algorithms perform equally can be rejected or not. Then, we conduct separate binary tests to compare the performances of each algorithm against each other. For the binary tests we adopt the \emph{Wilcoxon Signed-Ranks Test} method. For the multiple test we use the non-parametric \emph{Friedman Test}. In \cite{Demsar06Statistical}, Demsar recommends these tests as safe alternatives to the classical parametric t-tests to compare classifiers over multiple datasets.   

The main hypothesis of this paper is that our algorithms are faster than CVM. We have also observed than they are slightly less accurate. Therefore, our design for the binary tests between our algorithms and CVM is that of Tab. \ref{table:designTests}. As regards the comparison of the proposed methods, there is no an apparent advantage in terms of running time of one against the other. MFW seems however more accurate than FW. We thus conduct a two-tailed test for the running times but adopt a one-tailed test for testing accuracy. 

\begin{table}[h]
\centering
\begin{small}
\begin{tabular}{|p{1.6cm}|p{9cm}|}
  \hline
 & FW vs. CVM \\
     \hline
   Time & $H_0: \, \mbox{FW and CVM are equally fast} $\\
   &  $H_1: \, \mbox{FW is faster than CVM} \,$\\
   Accuracy & $H_0: \, \mbox{FW and CVM are equally accurate} $\\
     &  $H_1: \, \mbox{CVM is more accurate than FW}\,$\\
  \hline
  \end{tabular}
  \begin{tabular}{|p{1.6cm}|p{9cm}|}
  \hline
 & MFW vs. CVM \\
     \hline
   Time & $H_0: \, \mbox{MFW and CVM are equally fast} $\\
   &  $H_1: \, \mbox{MFW is faster than CVM} \,$\\
   Accuracy & $H_0: \, \mbox{MFW and CVM are equally accurate} $\\
     &  $H_1: \, \mbox{CVM is more accurate than MFW}\,$\\
  \hline
\end{tabular}
  \begin{tabular}{|p{1.6cm}|p{9cm}|}
  \hline
 & FW vs. MFW \\
     \hline
   Time & $H_0: \, \mbox{FW and MFW are equally fast} $\\
   &  $H_1: \, \mbox{Running times of MFW and FW are different} \,$\\
   Accuracy & $H_0: \, \mbox{FW and MFW are equally accurate}\, $\\
     &  $H_1: \, \mbox{FW is less accurate than MFW}\,$\\
  \hline
\end{tabular}\end{small}

\caption{\label{table:designTests} Null and alternative hypotheses for the binary statistical tests.}
\end{table}

In Tab. \ref{table:tests1} we report the values of the tests statistics calculated on the 26 datasets used in this paper. The critical values for rejection of the null hypothesis under a given significance level can be obtained in several books \cite{Mendenhall}. Here, in Tab. \ref{table:tests2}, we report the p-values
corresponding to each test.\footnote{For reproducibility concerns, p-values were computed using the statistical software R \cite{R}. For the Wilcoxon Signed-Ranks Test, the exact p-values were preferred to the asymptotic ones. The Pratt method to handle ties is employed by default. In the case of the Friedman test, the Iman and Davenport's correction was adopted, as suggested in \cite{Demsar06Statistical}.} 

\vspace{0.2cm}
\begin{table}[h]
\centering
\begin{small}
\begin{tabular}{|p{1.3cm}|p{2.2cm}|p{2.3cm}|p{2.2cm}|p{2.5cm}|}
  \hline
  & \multicolumn{3}{|c|}{W statistic } & F statistic  \\
 \cline{2-4}
 & FW vs. CVM & MFW vs. CVM & FW vs. MFW & FW,MFW,CVM \\
     \hline
   Time & 17 & 20 & 159 & 14.858 \\
   Accuracy & 19 & 48.5 & 63.5 & 11.879\\
  \hline
\end{tabular}\end{small}
\caption{\label{table:tests1} Values of the W and F statistics for Wilcoxon Signed-Ranks Tests and Friedman Tests respectively.}
\end{table}

\begin{table}[h]
\centering
\begin{small}
\begin{tabular}{|p{1.3cm}|p{2.2cm}|p{2.3cm}|p{2.2cm}|p{2.5cm}|}
  \hline
 & \multicolumn{3}{|c|}{Binary Tests} &Multiple Tests \\
 \cline{2-4}
 & FW vs. CVM & MFW vs. CVM & FW vs. MFW & FW,MFW,CVM \\
     \hline
   Time &  $3.085e-06$ & $5.528e-06$ & $0.6893$ & $1.72e-05$ \\
   Accuracy & $4.977e-06$ & $3.21e-04$ & $1.23e-03$ & $1.20e-04$\\
  \hline
\end{tabular}\end{small}
\caption{\label{table:tests2} P-values corresponding to the statistical tests.}
\end{table}

Note that in all but one case (binary test FW vs. MFW about running time) the p-values are lower than $0.01$. Therefore, for most commonly used significance levels ($0.01$, $0.05$, $0.1$, or lower) we conclude that there are significant differences in terms of time and accuracy among the algorithms. 
Table \ref{table:designTests} summarizes the conclusions from the binary tests. Note that the main hypothesis of this paper is confirmed. Most of the time our algorithms run faster and are less accurate. In the previous sections we have seen however that the loss in accuracy is usually lower than $1\%$, while the running time can be order of magnitudes better. As regards the comparison of the proposed algorithms FW and MFW, we cannot conclude that the difference in training time is statistically significant. However, we conclude that MFW is more accurate than FW. This last observation stresses the relevance of this work as an extension of the results presented in \cite{CIARP}.       
\vspace{0.2cm}
\begin{table}[h]
\centering
\begin{small}
\begin{tabular}{|p{1.6cm}|p{9cm}|}
  \hline
 & FW vs. CVM \\
     \hline
   Time &  $H_0$ rejected, so $H_1: \, \mbox{FW is faster than CVM} \,$\\
   Accuracy & $H_0$ rejected, so $H_1: \, \mbox{CVM is more accurate than FW}\,$\\
  \hline
  \end{tabular}
  \begin{tabular}{|p{1.6cm}|p{9cm}|}
  \hline
 & MFW vs. CVM \\
     \hline
   Time & $H_0$ rejected, so $H_1: \, \mbox{MFW is faster than CVM} \,$\\
   Accuracy & $H_0$ rejected, so $H_1: \, \mbox{CVM is more accurate than MFW}\,$\\
  \hline
\end{tabular}
  \begin{tabular}{|p{1.6cm}|p{9cm}|}
  \hline
 & FW vs. MFW \\
     \hline
   Time & We cannot reject $H_0: \, \mbox{FW and MFW are equally fast} $\\
   Accuracy & $H_0$ rejected, so $H_1: \, \mbox{FW is less accurate than MFW}\,$\\
  \hline
\end{tabular}\end{small}
\caption{\label{table:designTests} Conclusions from the binary statistical tests for significance levels $0.01$, $0.05$, $0.1$.}
\end{table}

\subsection{Experiments on the parameter $C$}
\label{subsec:C}

Previous experiments have shown that parameter $C$ used by SVMs to
handle noisy patterns can have a significant impact on the training
time required to build the classifier \cite{CIARP}. We hence conduct
experiments on some datasets to study this effect in more detail.

\begin{figure}[h!!!]
\centering
\begin{tabular}{c}
\includegraphics[width=\textwidth, height=0.22\textheight]{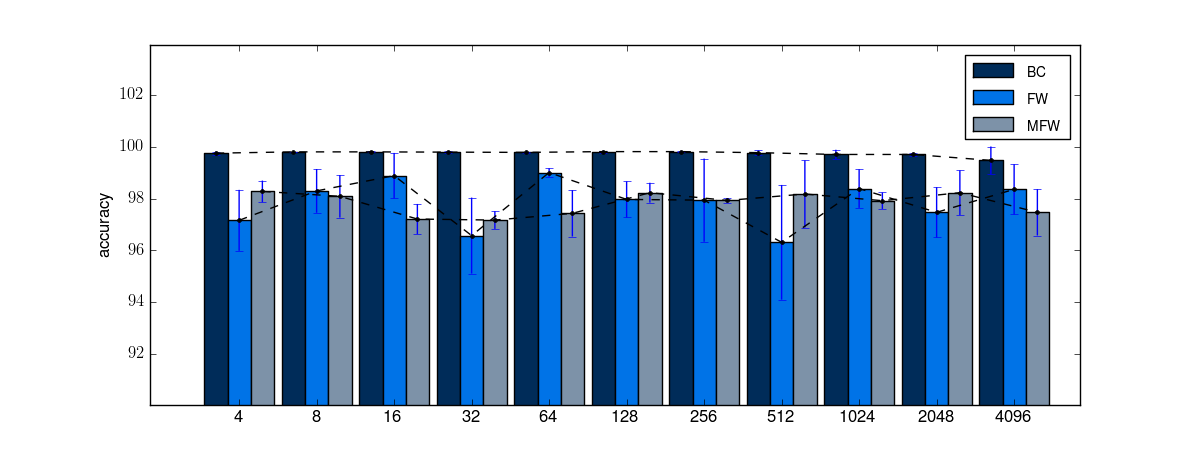}\\
\includegraphics[width=\textwidth, height=0.22\textheight]{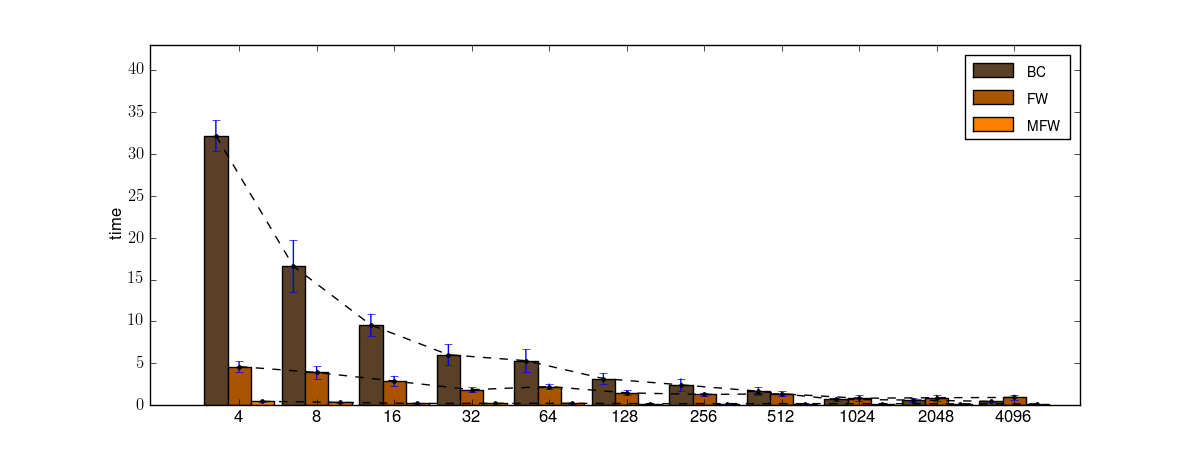}\\
\includegraphics[width=\textwidth, height=0.22\textheight]{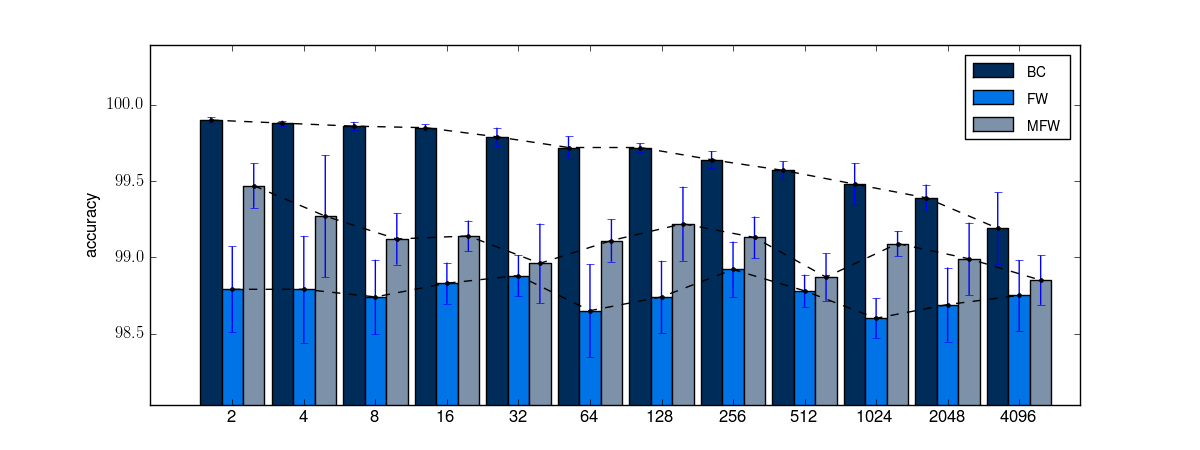}\\
\includegraphics[width=\textwidth, height=0.22\textheight]{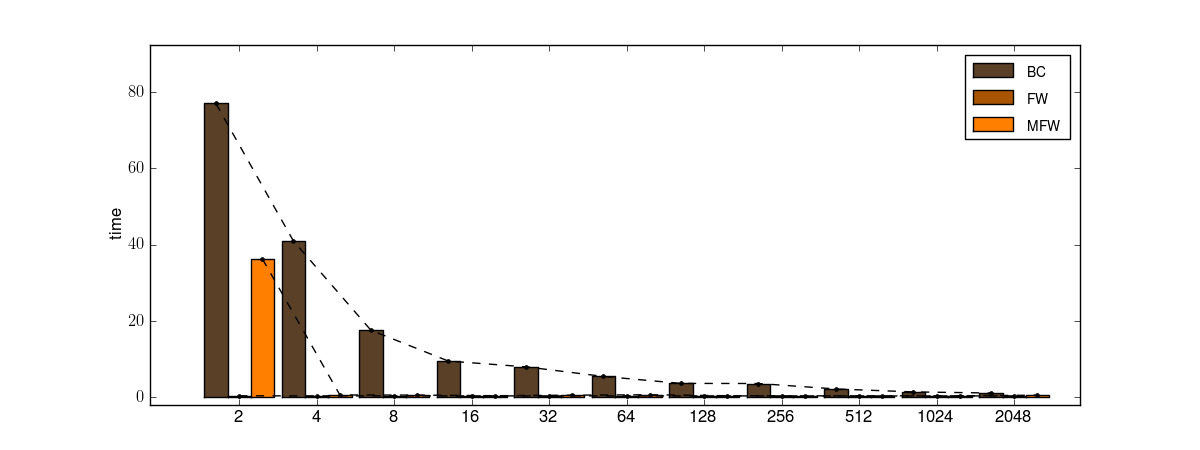}
\end{tabular}
\caption{Test accuracies (first row) and training times (second row)
obtained while changing the value of $C$ in the \textbf{Shuttle} and
\textbf{KDD99-10pc} datasets.}\label{fig:exp_C_shuttle}
\end{figure}

\begin{figure}[h!!!]
\centering
\begin{tabular}{c}
\includegraphics[width=\textwidth, height=0.22\textheight]{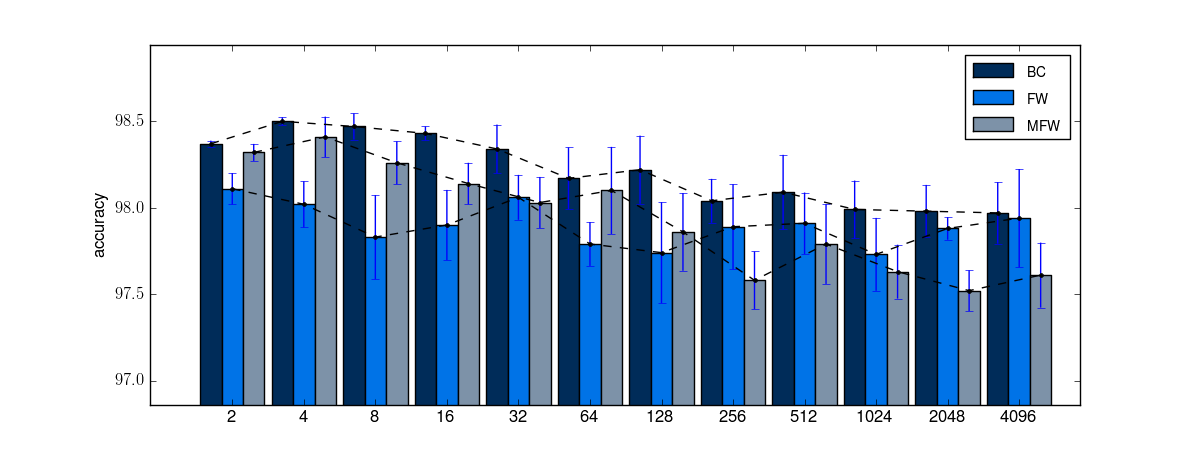}\\
\includegraphics[width=\textwidth, height=0.22\textheight]{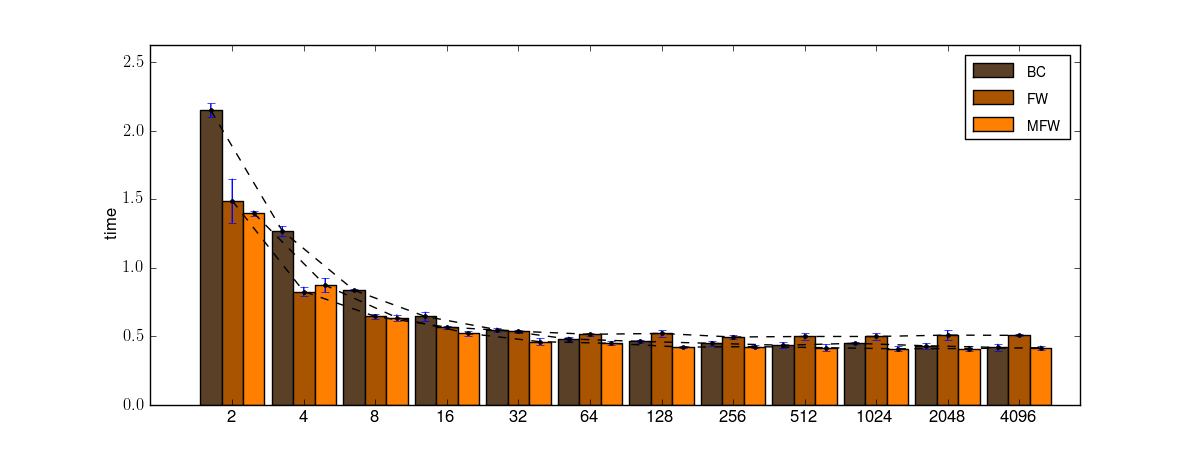}\\
\includegraphics[width=\textwidth, height=0.22\textheight]{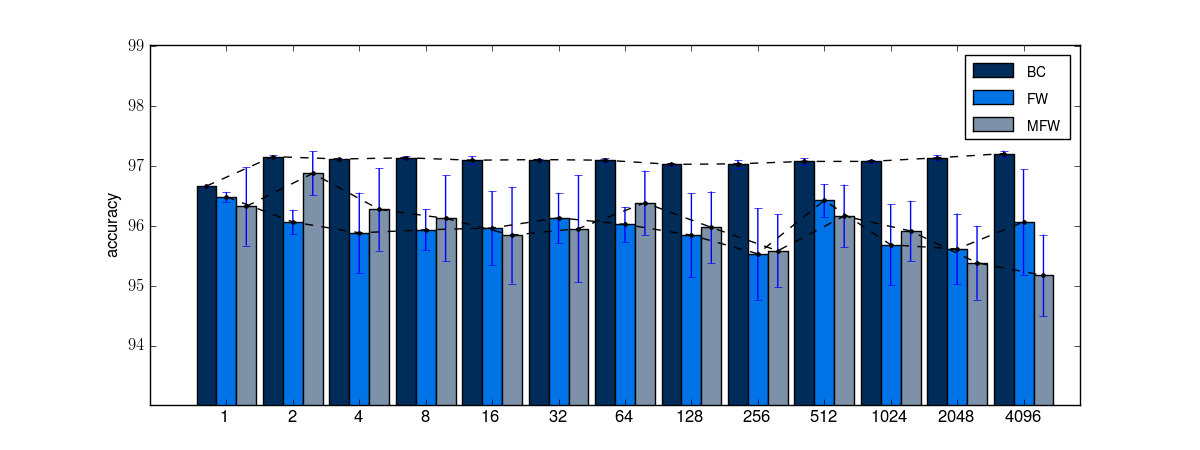}\\
\includegraphics[width=\textwidth, height=0.22\textheight]{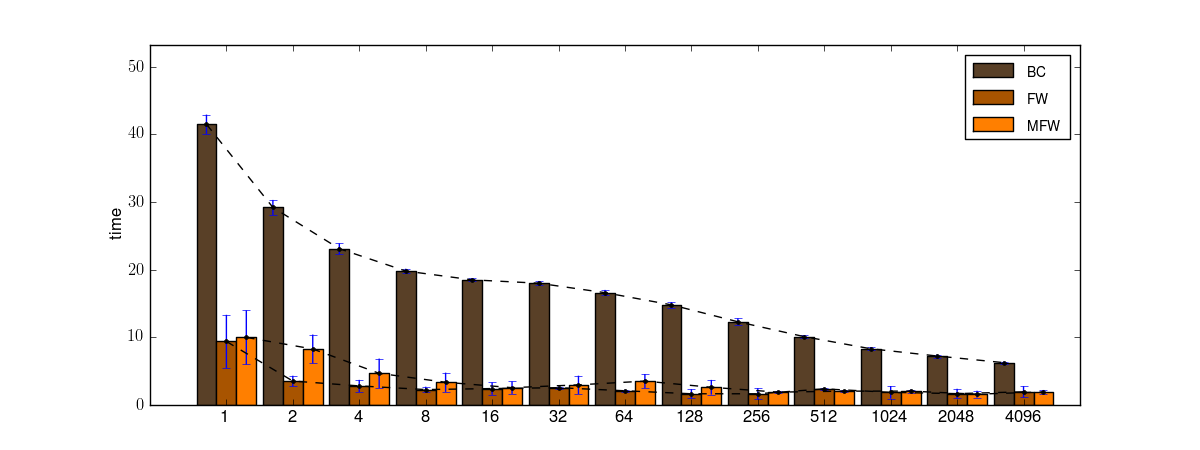}
\end{tabular}
\caption{Test accuracies (first row) and training times (second row)
obtained while changing the value of $C$ in the \textbf{Pendigits} and
\textbf{Reuters} datasets.}\label{fig:exp_C_pendigits}
\end{figure}

Figs. \ref{fig:exp_C_shuttle} and \ref{fig:exp_C_pendigits} show the
training times and accuracies obtained in the \textbf{Shuttle},
\textbf{KDD99-10pc}, \textbf{Pendigits} and \textbf{Reuters} datasets when
changing the value of $C$. Results confirm the general effect of
this parameter on the training time: as $C$ grows all the algorithms
become faster. However the training times of the proposed methods
are most of time significantly lower than those of CVMs,
independently of the value of parameter $C$ used by the SVM.

\cleardoublepage

\subsection{Experiments with Non-Normalized Kernels}
\label{subsec:otherKernels}
Solving a classification problem using SVMs requires to select a
kernel function. Since the optimal kernel for a given application
cannot be specified \emph{a priori}, the capability of a training method to
work with any (or the widest possible) family of kernels is an
important feature.

In order to show that the proposed methods can obtain
effective models even if the kernel does not satisfy the conditions
required by CVMs, we conduct experiments using the homogeneous second
order polynomial kernel $k(\bx_i,\bx_j)=(\gamma
\bx_i^{T}\bx_j)^{2}$. 
Here, parameter $\gamma$ is estimated as the
inverse of the average squared distance among training patterns.
Parameter $C$ is determined as usual by using a validation grid
search on the values $2^{0},2^{1},\ldots,2^{12}$. The test set is
never used to determine SVMs meta-parameters.

Note however that the purpose of this section is not to determine an optimal choice of the kernel function 
on the considered problems. The results presented below are merely indicative of the capability of the FW
and MFW methods to handle a wide family of kernels, thus allowing for a greater flexibility in building a classifier.

Tab. \ref{tab:results_poly} summarizes the results obtained in some
of the datasets used in this section. We can see that both test
accuracies and training times are comparable to those obtained using
the gaussian kernel. It should be noted that the CVM algorithm
cannot be used to train an SVM using the kernel selected for this
experiment and thus we only incorporate the FW and MFW methods in
the table. These results demonstrate the capability of our methods
to be used with kernels other than those satisfying the
normalization condition imposed by CVMs.

\begin{table}[ht]
\vspace{0.1cm}\centering
\begin{small}
\begin{tabular}{|@{\,\,\,\,}p{1.7cm}@{\,\,\,}|@{\,\,\,\,}p{1.6cm}@{\,\,\,}|@{\,\,\,\,}p{1.4cm} p{2.1cm} | @{\,\,\,\,} p{1.5cm} p{1.6cm}@{\,\,\,}|}
\hline
% & & & & & \\
Dataset & Algorithm & Accuracy & STD-Accuracy & Time(s) & STD-Time\\
% & & & & & \\
\hline
% & & & & & \\
 \textbf{Shuttle} &  FW &  96.58  &  9.44E-01  &  6.32E+01  &  3.69E+01 \\
 \textbf{Shuttle} &  MFW &  95.86  &  9.33E-01  &  8.52E+01  &  1.29E+02 \\
 \textbf{Reuters} &  FW &  95.80  &  3.00E-01  &  2.89E+00  &  3.23E-01 \\
 \textbf{Reuters} &  MFW &  95.90  &  1.98E-01  &  2.39E+00  &  1.52E-01 \\
 {\textbf{Web w1a}} &  FW &  97.22  &  1.08E-01  &  7.60E-02  &  3.50E-02 \\
 {\textbf{Web w1a}} &  MFW &  97.49  &  1.47E-01  &  2.52E-01  &  1.34E-01 \\
 {\textbf{Web w2a}} &  FW &  97.33  &  1.65E-01  &  1.42E-01  &  1.11E-01 \\
 {\textbf{Web w2a}} &  MFW &  97.09  &  1.93E-01  &  8.40E-02  &  6.53E-02 \\
 {\textbf{Web w3a}} &  FW &  97.32  &  1.68E-01  &  2.64E-01  &  1.33E-01 \\
 {\textbf{Web w3a}} &  MFW &  97.22  &  1.22E-01  &  3.18E-01  &  3.19E-01 \\
 {\textbf{Web w4a}} &  FW &  97.16  &  1.43E-01  &  3.76E-01  &  3.22E-01 \\
 {\textbf{Web w4a}} &  MFW &  97.25  &  1.49E-01  &  3.74E-01  &  3.65E-01 \\
 {\textbf{Web w5a}} &  FW &  97.08  &  7.37E-02  &  1.54E-01  &  2.14E-01 \\
 {\textbf{Web w5a}} &  MFW &  97.11  &  1.12E-01  &  2.78E-01  &  3.20E-01 \\
 {\textbf{Web w6a}} &  FW &  97.28  &  2.37E-01  &  2.96E-01  &  2.37E-01 \\
 {\textbf{Web w6a}} &  MFW &  97.18  &  1.31E-01  &  5.02E-01  &  4.11E-01 \\
 {\textbf{Web w7a}} &  FW &  97.23  &  6.22E-02  &  3.88E-01  &  2.81E-01 \\
 {\textbf{Web w7a}} &  MFW &  97.23  &  1.51E-01  &  2.10E-01  &  1.35E-01 \\
 {\textbf{Web w8a}} &  FW &  97.06  &  1.10E-01  &  2.76E-01  &  2.32E-01 \\
 {\textbf{Web w8a}} &  MFW &  97.24  &  2.91E-01  &  3.38E-01  &  3.32E-01 \\
%  & & & & & \\
\hline
\end{tabular} \end{small}
\vspace{0.2cm}
\caption{\label{tab:results_poly} Accuracies and training times obtained with a polynomial kernel. Statistics correspond to
the mean and standard deviation obtained from $5$ repetitions of each
experiment.}
\end{table}

\subsection{Discussion}
\label{subsec:discussion}

We now comment in more detail the results presented above. First of
all we note that, most of the time, the proposed algorithms appear
very competitive against CVM, with a tendency to favor training
speed in large datasets, sometimes at the expense of a little
accuracy. CVMs appear faster than FW just in three problems among the
single datasets studied in Subsection \ref{singledatasetsec}:
\textbf{Pendigits}, \textbf{USPS} and \textbf{Shuttle}. It should be
noted however that the \textbf{Pendigits} and \textbf{USPS} datasets
correspond to multi-category problems and are approached using a
decomposition method based on solving several binary subproblems. Now, as
shown in Tab. \ref{TabDatasets}, the greatest binary subproblem
for these datasets, is smaller than all the problems of the
\textbf{Web} collection and all but one of the \textbf{Adult}
collection. %The proposed methods have shown to scale better than
%CVMs on these datasets.
When each subproblem is very small, SMO iterations are quite cheap,
and the overall cost of running the BC procedure is reasonably low.
In these cases, training with a CVM (or even with a traditional SMO-based SVM)
possibly constitutes a convenient choice. The advantage of FW-based methods lies
instead in their capability to effectively handle larger problems, as results on the \textbf{Web} and \textbf{Adult}
collections show.

All the methods offer very similar testing performances on all the
character recognition problems (\textbf{Letter}, \textbf{Pendigits},
\textbf{MNIST}, \textbf{USPS-Scaled} and \textbf{USPS-Ext}). On
datasets \textbf{IJCNN} and \textbf{Reuters}, CVM offers more
accurate classifiers but employs a larger running time compared to
FW and MFW. The same can be said about the \textbf{KDD99-10pc}
problem, but in this case the speed-up offered by FW and MFW is
considerably larger, up to two orders of magnitude. The
\textbf{Shuttle} dataset returns mixed results, which are probably
due to the high lack of homogeneity in the size of the subproblems
solved in the OVO decomposition approach. Finally, the two FW
methods are clearly advantageous on \textbf{Protein} and
\textbf{KDD99-Full} datasets, where they offer the same accuracy of
CVMs along with a considerably improved running time.

The results on the \textbf{Web} and \textbf{Adult} datasets are of
particular interest and deserve further comment. They consist of a
series of datasets of increasing size, and from their study we
expect to gain an understanding of the performance of the algorithm
as $m$ gradually increases. In fact, as documented in 
\cite{platt99smo-seminal} and \cite{MM99} these datasets have
been commonly used to compare the scalability of SVM algorithms. In
this regard, our results appear very encouraging. Not only do both
FW algorithms outperform CVM in every instance of the \textbf{Web}
collection with respect to CPU time, but the observed speedup
increases monotonically as the dataset size increases, reaching a
peak of two orders of magnitude for the FW method. Both algorithms
also outperforms running times of CVM on all but two datasets of the
\textbf{Adult} collection, with very similar testing accuracies.

The clear advantage of the MFW method with respect to both FW and
CVM in the \textbf{Adult} series can be probably explained by the
considerable size of the support vector set, which roughly amounts
to a $60\%$ of the full dataset, for all the methods. It is evident
that, if $\mathcal{C}_k$ becomes large, SMO iterations become quite
expensive, slowing down the CVM procedure. As regards the advantage
of MFW over FW, we interpret the results as follows. At the
beginning of the training process, the algorithms start with a small
approximating ball, and progressively expand it by including new
examples. Intuitively, in the first iterations both methods tend to
include a large number of points in order to increase the radius of
the ball (and thus the objective value). Some of these examples do
not belong to the optimal support vector set and the algorithms will
try to remove them from the model once they approach the solution.
When the support vector set is large, as in this case, the number of
these spurious examples can be quite large, hampering the progress
towards the optimum. Now, FW is not endowed with the possibility of
explicitly removing points from the current coreset approximation,
implying that the weights of useless patterns vanish only in the
limit. That is, a large number of iterations may be taken before
they drop below the tolerance under which they are numerically
considered zeros. MFW, in contrast, possesses the ability to remove
undesired points directly, and thus enjoys a considerable advantage
when the number of such examples is not small.

\section{Conclusions and Perspectives}

In this paper we have described the application of $\epsilon$--coreset based methods from computational geometry to the task of efficiently training an SVM, an idea first proposed in \cite{coreSVMs05tsang}. We have introduced two algorithms falling in this category, both based on the Frank-Wolfe optimization scheme. These methods use analytical formulas to learn the classifier from the training set and thus do not require the solution of nested optimization subproblems. Compared with the results we presented in \cite{CIARP}, we have explored a variant of the algorithm which compares favorably in terms of testing accuracy and achieves training times similar to our original version. 

The large set of experiments we report in this paper confirms and considerably expands the conclusions reached in \cite{CIARP}. As long as a minor loss in accuracy is acceptable, both Frank-Wolfe based methods are able to build SVM classifiers in a considerably smaller time compared to CVMs, which
in turn have been proven in \cite{coreSVMs05tsang} to be faster than most traditional SVM software. These conclusions were statistically assessed using non-parametric tests. A second contribution of this work has been to present preliminary evidence about the capability to handle a wider family of kernels than CVMs, thus allowing for a greater flexibility in building a classifier. Further variations of this procedure will be explored in a future work, including learning tasks other than classification.

\bibliographystyle{abbrv}
\bibliography{IJPRAI11-bibliography}

\end{document}